\documentclass[10pt,onecolumn]{IEEEtran}
\usepackage{geometry}
\usepackage[switch]{lineno}
\usepackage[cmex10]{amsmath}
\usepackage{amssymb,latexsym,amsfonts,stmaryrd}
\usepackage{amsthm}
\usepackage{mathtools}
\usepackage{graphicx}
\usepackage[noadjust]{cite}
\usepackage{algorithm}
\usepackage{algorithmic}
\usepackage{subfigure}
\usepackage{color}
\usepackage{soul}
\usepackage{lineno}
\usepackage{mathptmx}
\usepackage[ruled,vlined,algo2e]{algorithm2e}
\usepackage{longtable}
\usepackage{etoolbox}
\usepackage{adjustbox}
\usepackage{multirow}
\makeatletter
\patchcmd{\@makecaption}
  {\scshape}
  {}
  {}
  {}
\makeatother

\usepackage[hyphens]{url}
\usepackage[hidelinks]{hyperref}
\hypersetup{breaklinks=true}
\usepackage{cite}

\usepackage{natbib}
\usepackage{cleveref}

\setcitestyle{authoryear}
 
\title{Worldwide city transport typology prediction with sentence-BERT based supervised learning via Wikipedia}

\author{
  \IEEEauthorblockN{Srushti Rath and  Joseph Y.J. Chow \\}
    Department of Civil and Urban Engineering\\
    New York University, NY, USA  \\
    Email: srushti.rath@nyu.edu,  joseph.chow@nyu.edu}
\date{May 2021}

\begin{document}
\maketitle
\section*{Abstract}
An overwhelming majority of the world's human population lives in urban areas and cities.
Understanding a city's transportation typology is immensely valuable for planners and policy makers whose decisions can potentially impact millions of city residents. Despite the value of understanding a city's typology, labeled data (city and it's typology) is scarce, and spans at most a few hundred cities in the current transportation literature. To break this barrier, we propose a supervised machine learning approach to predict a city's typology given the information in its Wikipedia page. Our method leverages recent breakthroughs in natural language processing, namely sentence-BERT, and shows how the text-based information from Wikipedia can be effectively used as a data source for city typology prediction tasks that can be applied to over 2000 cities worldwide. We propose a novel method for low-dimensional city representation using a city's Wikipedia page, which makes supervised learning of city typology labels tractable even with a few hundred labeled samples. These features are
used with labeled city samples to train binary classifiers (logistic regression) for four different city typologies: (i) congestion, (ii) auto-heavy, (iii) transit-heavy, and (iv) bike-friendly cities resulting in 
reasonably high AUC scores of 0.87, 0.86, 0.61 and 0.94 respectively. Our approach provides sufficient flexibility for incorporating additional variables in the city typology models and can be applied to study other city typologies as well. Our findings can assist a diverse group of stakeholders in transportation and urban planning fields, and opens up new opportunities for using text-based information from Wikipedia (or similar platforms) as data sources in such fields.\\

\noindent Keywords: city transport typology, SBERT, global cities, Wikipedia\\

\noindent \textit{*Preprint to appear in Transportation Research Part C}

%% INTRODUCTION %%
\section{Introduction} \label{sec:introduction}
By 2050, the share of the people residing in urban areas is projected to reach 70\% of the global population (\citealp{UNreport}).
Rapid urban expansion; fast-evolving transportation trends especially in areas of autonomy, electrification and connectivity; and such new business models as shared-use mobility-on-demand and Mobility-as-a-Service (\citealp{pantelidis2020many}) are influencing urban policymakers to improve mobility systems around the world. The way these mobility solutions evolve across different cities vary in scale and pace based on city characteristics. 

In today's global mobility market, mobility operators can learn much from the outcomes of transportation solutions in one city for deployments and planning in other cities owing to commonalities in residents' perception in various aspects (\citealp{grauwincommon,Mckinsey}). The effectiveness with which one manages their city deployment portfolio can make or break the service. For example, electric bikes were launched in several select cities like Austin, Portland, and Baltimore (\citealp{austinwheels}; \citealp{lee2019forecasting}). 
Today, electric scooters operate in a large number of cities around the U.S.  (\citealp{BTSescooters}). Meanwhile, Chariot (an on-demand shuttle service that was acquired by Ford) started its microtransit services in San Francisco in 2014 %(\citealp{austinchariot}) 
and subsequently expanded its operations to Austin, Seattle, %(\citealp{seattlechariot}) 
and a few other cities. However, due to low adoption of their services among riders, Ford shut down its operations (\citealp{ford}). Hence, understanding similarities in cities and their typologies can be a key factor in facilitating the planning, assessment and evaluation of various mobility solutions at the city level. 

\begin{figure}[!htb]
\begin{center}
\includegraphics[scale=.32]{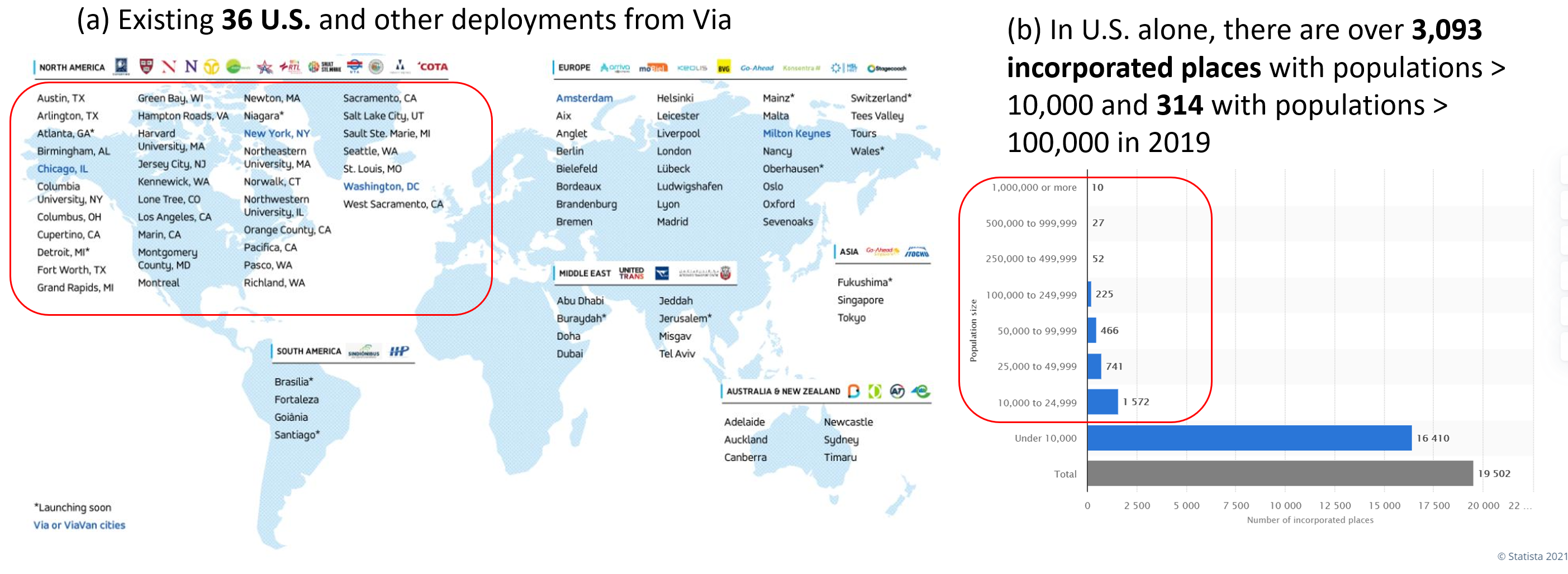}
\caption{(a) Via deployments around the world including 36 cities in the U.S. (cooee.busways.com.au/the-ponds/via); (b) distribution of incorporated places in the U.S. (\citealp{statista})} 
\label{fig:Viadeployments} 
\end{center}
\end{figure}

The task is nontrivial. Consider for example the microtransit company Via. As shown in Figure~\ref{fig:Viadeployments}(a), they have a substantial deployment portfolio around the world including 36 cities in the U.S. Nonetheless, there are over 314 cities in the U.S. with population over 100K and over 3,093 incorporated places with populations over 10K. \textcolor{black}{Knowledge on city typologies can be useful to mobility operators in the service deployment decision process (e.g., identifying target markets/cities to enter), as part of a complex process that involves agency engagement, pilot tests, marketing studies, etc. Ultimately, operators would have a limited set of deployment data from which they seek to forecast performance measures (e.g., last mile ridership, vehicle miles traveled). However, cities like Columbus, Austin, and Sacramento can have very different behavioral characteristics. Having typologies would make it possible to segment the city data to have more relevant forecasts (\citealp{Viastudy}). The typologies can be especially beneficial for analyzing or estimating future impacts of certain service operations, strategies, or policies in cities when considering the emerging transportation aspect of services like microtransit. This is because even though each city may have their individual demand supply interaction characteristics, it is not feasible to model or simulate each new city of interest in the world to conduct such analysis; limited data on such new technology deployments makes this even more difficult. This is the same motivation for the whole field of city typology classification, as exemplified by \cite{oke2020evaluating}'s use of typologies to study auto-dependent prototype cities.}

City typologies have been proposed in the literature based on broad economic and geographic forms. For example, \cite{harris1943functional} classified cities by economic functions like manufacturing, retail, education, etc., while  \cite{creutzig2015global} proposed 8 typologies oriented around socioeconomic and environmental indicators to classify 274 cities. In terms of transportation metrics, studies have found that cities do exhibit commonalities, whether it is in road networks (\citealp{louf}) or public transit services (\citealp{derrible2010complexity, fielbaum}). A number of research studies have also identified typologies (\citealp{thomson, priester, okeMIT}) for these cities. \cite{okeMIT} present a hierarchical typology framework to classify 331 cities worldwide into transportation-based typologies; a brief high-level description of the six city typologies in the highest level of their hierarchical typologization is listed in Table~\ref{tab:mit_typologies}.
\begin{table}[h]
    \centering
    \caption{Summary of the high-level city typologies based on \cite{okeMIT}}
    \begin{tabular}{|l|c|c|}
    \hline
    City typology & Description & Example cities\\
     \hline
        Auto      &  Auto-dependent & Washington DC,  Toronto,
        Raleigh, Kuwait City \\
        Bus transit     &  High usage of bus transit  & Rio de Janeiro, Jakarta, Tehran, Mecca \\
        Congested & Congestion in cities &  Bangalore,   Lagos, Manila, Port-au-Prince\\
         Metro bike & High bike share and metro & Ningbo, Zhengzhou, 
        Shenzhen, Chongqing\\
         Mass transit & High usage of mass transit & Singapore, Seoul, Tel Aviv,  London  \\
         Hybrid & Mix of mode choices & Busan, Lisbon,   Santiago, Johannesburg\\
    \hline
    \end{tabular}
    \label{tab:mit_typologies}
\end{table}

The main research gap in the literature lies in the data availability. Although prior studies on mobility-oriented city classification or typologization have successfully identified city typologies, they relied mainly on datasets made available by city agencies, private sector companies, transportation operators, and universities. As such, application of these typologies to cities at a large, global-wide scale, e.g., on the order of thousands of incorporated places around the world, is not currently feasible, especially across multiple countries.  

While a growing number of cities worldwide are recognizing the potential of open sourcing data for decision-making and planning purposes, many cities have yet to embrace an open data approach (\citealp{torrebig}). Lately, new sources of data that are generated continuously by users (including data from social media platforms, geo-positioning system (GPS), smart cards, smartphones and other web-based sources) have gathered a global interest in the transportation industry and urban studies (\citealp{thakuriah, rashidi, hu, chow2018informed}). However, if the data of interest in such platforms are beyond geo-locations, check-ins, hash-tags, sentiments and keywords, extracting meaningful information from long and unstructured text can be challenging. 

We explore the usage of Wikipedia as such a source of data for identifying city typologies. Wikipedia is unique in many aspects; it is essentially the largest digital encyclopedia worldwide that is powered by millions of crowd-sourced content editors and moderators. Enhanced at a rate of over 1.9 edits every second (\cite{wikistats}), Wikipedia serves as a reliable and inexpensive source of information.
Wikipedia not only provides a wide-range of information on various aspects of cities (such as transportation, demography, geography, economy, environment, education, culture, and others) in a consolidated manner, but also does it at an unparalleled scale (\emph{i.e.}, covering thousands of cities across the world for which detailed data may not be available otherwise). 

\begin{figure}[!htb]
\begin{center}
\includegraphics[scale=.51]{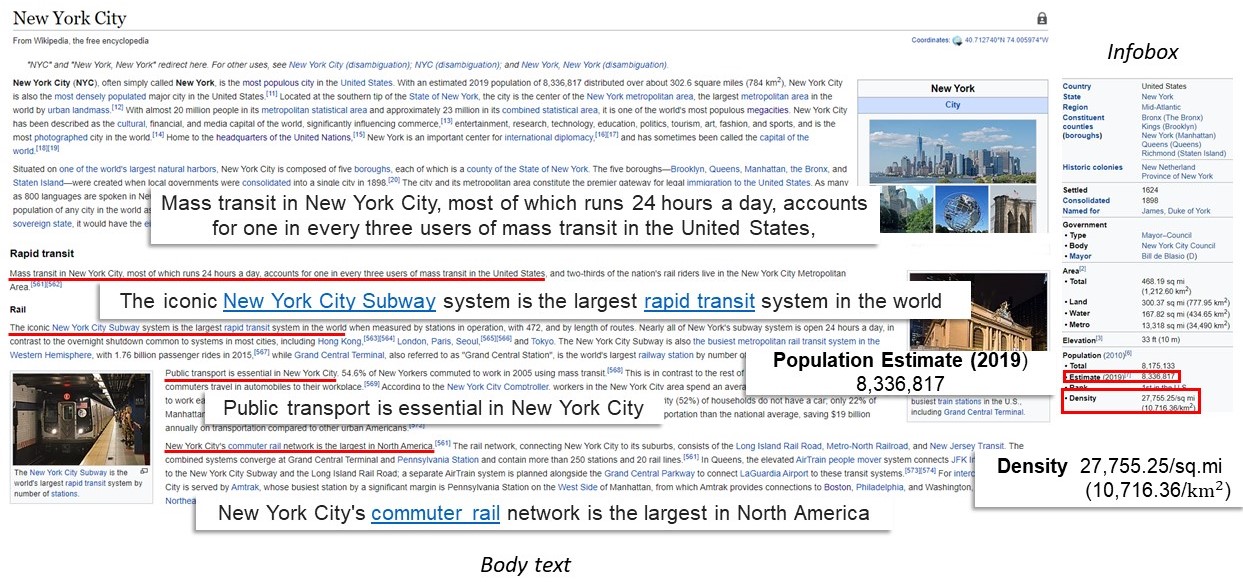} 
\caption{New York City's Wikipedia page: The infobox is structured with fields and their values, whereas the article body text is unstructured. Lines indicative of the city typology (transit-heavy as per \cite{okeMIT}) have been highlighted.} \label{fig:wikipage_structure} 
\end{center}
\end{figure}

For example, in a recent study  by \cite{okeMIT}, New York City (NYC) was assigned the \textit{transit-heavy} city typology label based on multiple data sources excluding Wikipedia. Assuming, the Wikipedia page on NYC has supporting evidence that the city is indeed transit-heavy, if a human reader was tasked with identifying supporting lines,
it would be an easy task to surface the lines highlighted in Figure~\ref{fig:wikipage_structure}. (which strongly support the transit-heavy label).
Furthermore, the Wikipedia page may also contain information on different aspects of the city, including recent demographic estimates and numbers from the infobox fields, which may be latent factors influencing the typology.
Similar information from Wikipedia on transportation scenarios in other cities could plausibly paint a picture of the typology of respective cities.
But extracting such information in an automatic manner, \emph{e.g.}, via a model which \textit{understands} the city's Wikipedia page from the perspective typology prediction is a challenging task. 

In an effort to explore the utility of large scale (and fresh) Wikipedia data, we broadly consider the task of a city's typology (category) prediction using information in the city's Wikipedia page. In spirit, this study also reinforces ongoing efforts in the transportation planning community on utilizing crowd-sourced information for urban planning and decision-making (\citealp{martisocial,ilieva, hasan}). 

The textual component in Wikipedia pages on cities is mostly in the form of such long and unstructured text. To illustrate this complexity, Figure~\ref{fig:wikipage_structure} shows sections of the Wikipedia page on NYC. As can be seen, the page mostly has unstructured text (with multiple paragraphs), and supporting images. It also has an infobox (on the right corner of the page), which has structured data (\emph{e.g.}, the city's population, and area).

Due to challenges in extracting useful information from long unstructured text, Wikipedia has remained a largely unexplored data source for transportation related studies. In the larger context of text understanding, the task of understanding and representing sentences (and paragraphs) for downstream prediction tasks has been an area of active research in the natural language processing (NLP) research community for several decades. Initial methods were based on word counts (\emph{e.g.}, term frequency-inverse document frequency (TF-IDF) \citealp{jones,salton}) where a sentence (or a collection of sentences forming a document) was simply represented by weighted occurrence count of each word in a given vocabulary.

However, such representations suffered from high dimensional vectors (length equal to size of the vocabulary, typically in the order tens of thousands for English language), and lack of semantic understanding (\emph{e.g.}, two words like 'auto' and 'car' will be considered widely different although they are similar in meaning). Over the years, researchers developed low-dimensional sentence embedding methods like Doc2Vec (\citealp{doc2vec}), and more recently sentence-BERT (\citealp{sbert}), which can map a sentence to a low dimensional vector (typically around $300$ dimensions for Doc2Vec, and $768$ for sentence-BERT) while preserving the meaning of the sentence. In particular, sentence-BERT is derived from a deep learning model called Bidirectional  Encoder  Representations  from  Transformers (BERT) (\citealp{bert}) which has revolutionized NLP applications over the last couple of years by capturing the meaning of a sentence at an unprecedented quality. For example, embeddings based on BERT (and derivatives like sentence-BERT) have achieved state-of-the-art performance for semantic textual similarity (\emph{e.g.}, comparing if two sentences are similar in meaning), sentiment analysis, and many other text classification tasks (\citealp{bert, sbert}).

Inspired by the success of sentence-BERT in popular NLP tasks, we build on top of it to extract useful information from Wikipedia pages in the context of city typology prediction.
However, having a sophisticated text understanding model like sentence-BERT (also called SBERT) does not trivially solve the problem of city typology prediction using Wikipedia pages. The challenge lies in:
(i) having very few labeled samples (\emph{i.e.}, few hundreds in existing datasets on cities with typology labels), coupled with the fact that
(ii) SBERT embeddings are of size $768$ (exceeding the number of labeled samples). In other words, training a supervised model which directly learns weights for SBERT embeddings (from the lines in a Wikipedia page) to predict a city typology label is not feasible. To address this major challenge, and to maintain interpretability, we propose a low-dimensional representation of a city's Wikipedia page as outlined below.

We propose a novel method to algorithmically extract lines from a city's Wikipedia page which semantically match a known set of possible typologies (\emph{e.g.}, congested, transit-heavy, auto-heavy or bike-friendly), 
and use the typology-wise match scores to form a $4$-dimensional vector representation (feature vector) for a city's Wikipedia page.
In addition, we use information from structured components like the infobox (\emph{e.g.}, population density) as an additional numeric feature for a city, and use the resultant low-dimensional feature vector for training a logistic regression model for city typology prediction. In particular, we use the labels in \cite{okeMIT} as ground truth labels for $\sim 300$ cities, and we adopt a one-versus-all approach for multi-class classification (\emph{i.e.}, train binary classifiers for $4$ different city typologies, and study their prediction accuracy); the justification behind this approach is provided in Section~\ref{sec:objective}.
With such trained models, we can easily propagate the labels in $\sim 300$ cities in \cite{okeMIT}, to over $2,000$ cities in Wikipedia.

Our main contributions can be listed as follows:
\begin{enumerate}
    \item We propose a low-dimensional representation of a city's Wikipedia page for the task of city typology prediction. The representation is based on algorithmically identifying lines in the Wikipedia page which semantically match (via SBERT) a known typology, and use their match scores to form features.
    Our approach combines both textual  and numeric data (from infobox) in Wikipedia pages, while still keeping the feature vector size low enough for supervised training with $\sim 300$ labeled samples.
    We achieve a binary classification AUC of 0.87, 0.86, 0.61, and 0.94 for our classifiers for congestion, auto-heavy, transit-heavy, and bike-friendly cities respectively, thereby demonstrating the efficacy of our proposed approach.
    \item We propose an iterative \textit{keyline} expansion method which finds a set of representative lines (which we refer to as keylines) from Wikipedia pages of cities; these representative (key)lines allude to a known city typology, and are crucial for identifying similar lines in cities beyond the training data set. In addition, the keyline expansion method is interpretable, and offers valuable insight into how Wikipedia content contributors express whether a city is congested, auto-heavy, transit-heavy or bike friendly.
    \item Using our trained model, we predict city typology scores (for different typologies) for over 2,000 cities present in Wikipedia. To the best of our knowledge, this is the largest dataset/analyses in scope on mobility-based city typology inferences. We also share valuable insights on how the proposed methods can go beyond typology prediction (\emph{e.g.}, estimate the chances of adopting microtransit services in cities).
\end{enumerate}

The paper is organized as follows. In Section~\ref{sec:related work}, we provide a
brief literature review, Section~\ref{sec:problem formulation} describes the mathematical formulation for the problem and discusses the objective. Section~\ref{sec:methodology} explains the proposed method and the algorithmic techniques that we employ in our study; this is followed by experiments in Section~\ref{sec:experiments} with results and discussion in Section~\ref{sec:results}. We conclude in Section~\ref{sec:conclusion}.

%%%% RELATED WORK %%%%%
\section{Related Work} \label{sec:related work}
\subsection{City typology understanding}
City typologies or profiles based on the dynamics of mobility in cities can allow easy identification of comparable cities for learning best practices and policies in the urban mobility planning context. Various studies can be found that are oriented towards understanding city patterns and urban forms. These studies primarily consider social and economic factors (\citealp{harris,bruce,martin}), geography or spatial metrics (\citealp{anas,sun,boeing2019},  and urban forms (\citealp{conzen, fielbaum,tsai,kasanko}). In the pursuit of developing approaches for interpreting global cities, \cite{huang2007global} utilize satellite images to classify 77 metropolitan areas worldwide by the urban form. \cite{louf} categorize 131 cities across the world based on the street patterns. Focusing on the expansion patterns in urban areas, \cite{kuang} conducts a comparative analysis of megacities in China and the U.S. In this direction, only a few studies have focused on the transportation aspect of cities. An analysis of 30 megacities by \cite{thomson} yields five different metropolitan archetypes based on transportation structures in cities. \cite{cervero} classify cities based on land use patterns and the structure of the transit system. \cite{priester} cluster 41 megacities worldwide into seven distinct types namely:  auto, non-motorized, hybrid, paratransit, transit, traffic-saturated and
the singleton (Manila). A recent study by \cite{okeMIT} uses hierarchical clustering to present a mobility-based typologization covering 331 cities (across the globe) using factors related to transportation, demographic, geographic, economic, and environmental dimensions of cities. In their study, authors present 12 typologies grouped into 6 high level categories (refer Table~\ref{tab:mit_typologies}). For the most part, researchers have restricted the scope of their analysis to a limited number of cities due to data scarcity in many cities (especially in developing countries). 

\subsection{Usage of crowd-sourced data in transportation and urban planning}
The advent of user-generated information (including data from social media accounts, GPS, smart cards, and mobile phones) and the ease of data access opened up various opportunities for multidisciplinary research. In the context of city understanding (particularly focusing on the transportation aspect), several studies have utilized location-based user-generated data (\emph{e.g.}, social media check-ins, geo-tags, GPS, and smartphone data) (\citealp{zhan}). In this setting, the Livelihoods Project by \cite{cranshaw} classifies the livelihood dispersion patterns in a city using geo-location data. \cite{hasan} characterize human patterns in a city based on purpose-specific activities using location-based social media data. Similarly, \cite{louail} study the morphological patterns in 31 Spanish cities. \cite{lenormand} perform a systematic comparison between five cities in Spain based on the land use patterns from mobile phone records.  \cite{calafiore} model cities as series of global urban networks to obtain functional neighbourhoods based on human dynamics and their contexts, across a sample of 10 global cities. A detailed overview of big data analysis for the systematic study of cities and urban phenomena can be found in \cite{lenormand2016} and \cite{martisocial}. 

\subsection{Text understanding}
Besides the use of geo-locations, keywords, and sentiments retrieved from user-generated platforms, mining data from the textual content has received growing attention in the past few years (\citealp{camacho}), since they contain latent information about people's views and experiences. Such information harvested from the crowd-sourced data can be of a great value in transportation related studies (\citealp{gal}).
However, automatically deriving useful information from informal and unstructured textual data can be very challenging. Furthermore, when going from a single line of text to a long paragraph (document) of text, the complexity of understanding increases substantially due to the presence of multiple interconnected references in a document (\citealp{doc2vec}). 

Depending on the availability or unavailability of labels (annotations) for textual data, the downstream text understanding models can be either supervised or unsupervised. In the realm of unsupervised text understanding models, topic models (\emph{e.g.}, Latent Dirichlet Allocation (\citealp{lda})) have been used for classifying regions based on their functions, and characterizing activity patterns in cities using points-of-interest and geo-located mobility data (\citealp{gao2017extracting,yuan,gal, hasan2014}).
In comparison, we focus on supervised approaches for text (document) understanding given typology labels for cities and their corresponding Wikipedia pages (documents). We describe below relevant background on Wikipedia, and techniques for supervised text understanding.

Wikipedia is a rich source of crowd-sourced information on cities. The information in Wikipedia is generated at no cost (updated by numerous contributors worldwide), and verified regularly by moderators. So far, it has been a free online service, and is also freely available for off-line analysis. However, much of the useful (qualitative) information is in a textual format (in unstructured article bodies), and extracting such information automatically is difficult. A few recent studies have looked into Wikipedia as a potential indicator of city characteristics (\emph{e.g.}, smart city related expressions (\citealp{cronemberger}), and for understanding poverty and education across sub-Saharan African nations (\citealp{sheehan})). However, to the best of our knowledge, there is no prior work on predicting a city's typology using its Wikipedia page; we believe that the hardness in understanding unstructured Wikipedia text, and limited labeled data on city typologies (\citealp{okeMIT}) may have contributed to the lack of prior work in this direction. We propose an approach for Wikipedia text understanding for city typology prediction; our approach is not only designed for low volume of labeled data, but is also interpretable in nature. Relevant background for our proposed text understanding approach is described below.

Fundamentally, in our supervised learning setup, the objective is to map a Wikipedia page (on a city) to a label (\emph{e.g.}, binary label indicating whether it is transit-heavy or not). However, in its raw form, the Wikipedia page is a collection of lines (with multiple words in a line). To represent such a page numerically (for training a machine learning model and eventually mapping it to a label), one of the earliest approaches was to simply encode it in a \textit{bag-of-words} fashion, \emph{i.e.}, by using a vector of size equal to the vocabulary, and populating it with weighted word counts (TF-IDF). However, such an approach led to high feature dimensions (English documents can easily lead to a vocabulary size in the order of tens of thousands), and the bag-of-words representation did not capture the sequencing of words in a document (the sequence can easily alter the semantics). 

A subsequent approach, Doc2Vec, made progress towards low-dimensional document representation (embedding), but was still limited in its ability to capture sequences and understand context. Around the same time, for short text (\emph{e.g.}, a sentence), using the average of low-dimensional word embeddings (of words in the sentence) also became a popular method for sentence representation; Word2Vec (\citealp{word2vec}), FastText (\citealp{fast_text}), and GloVe (\citealp{glove}) are examples of such low-dimensional word embedding methods. However, since the average does not capture the exact sequence of words, such methods are still limited in capturing the semantics.

A major breakthrough in the sentence representation problem came with the introduction of BERT which uses a neural network architecture called Transformer (\citealp{vaswani_transformers}) to capture the exact sequence of words and learn the context (by utilizing both the forward sequence and the backward sequence, and hence becoming bidirectional in nature). BERT and its variants for text representation led to the state-of-the-art results for many NLP tasks (\emph{e.g.}, text classification, and sentiment analysis). Furthermore, sentence-BERT (\citealp{sbert}) leveraged a pre-trained BERT model (trained on millions of English examples), and was fine tuned for textual similarity. In other words, sentence-BERT can be used to get a $768$-dimensional embedding for a sentence, and the semantic similarity of two sentences can be effectively gauged by the cosine similarity between their embeddings. We leverage this property of sentence-BERT embeddings in our proposed approach.

While BERT based methods have led to state-of-the-art results in standard natural language processing tasks, they are still relatively new (since 2019) and have not been leveraged yet in the transportation research community. Prior works using text understanding techniques for transportation related applications include only the usage of older methods like TF-IDF, Word2Vec, and FastText word embeddings. In \cite{kuflik}, the authors explore a TF-IDF based approach to extract and analyze transport related social media content. They highlight the challenges of extracting useful transportation related information from user-generated content using bag-of-words approach and note the potential of advanced text understanding techniques for efficiently mining information from crowd-sourced data. Another study by \cite{bencke} uses TF-IDF for automatically classifying social network messages relating to smart cities services (including transportation among other aspects). \cite{yao} integrate point-of-interest data and Word2Vec for studying the spatial distribution of urban land use.
A recent study by \cite{bondielli} uses FastText to categorize city areas based on tags associated with online news articles.

Finally, it should be noted that simply using sentence-BERT does not solve the problem of city typology prediction using Wikipedia; the remaining bottleneck comes from the low volume of labeled data (even lower than $768$, which is the sentence-BERT dimension size), and the restriction that sentence-BERT can only process a sentence and not Wikipedia pages comprising of hundreds of lines. As described later, we propose a novel method to leverage sentence-BERT embeddings of lines in a Wikipedia page, to form a low-dimensional representation of the page such that supervised training with a few hundred labeled cities is sufficient.

%%%% PROBLEM FORMULATION %%%%%
\section{Problem formulation} \label{sec:problem formulation}
In this section, we formally define the city (typology) label prediction problem using Wikipedia data. Section~\ref{sec:setup} provides a formal description of the setup with the notations used in this paper, and Section~\ref{sec:objective} explains the objectives of the predictive models trained on Wikipedia pages in our setup.

%% SETUP %%
\subsection{Setup} \label{sec:setup}
%% WIKIPEDIA DATA %%
\subsubsection{Wikipedia data} \label{sec:wikipedia data}
We assume a set $\mathcal{W}$ of Wikipedia pages (size of the set denoted by $|\mathcal{W}|$). Each page $W_i \in \mathcal{W}$ corresponds to a unique city (anywhere in the globe). We focus on two components of each page in  $\mathcal{W}$: (i) unstructured text from different sections (main body), and (ii) structured data from the infobox (like the one illustrated in Figure~\ref{fig:wikipage_structure}). Sections commonly found in city Wikipedia pages include demography, geography, history, economy, education, and transportation (not necessarily with the same section titles), along with a general description of the city in the introduction paragraph. In our setup, for simplicity, we ignore text from the footnotes and references, as their mentions in the main body are preserved and typically provide enough context. In addition, from the infobox, we extract demographic information on population densities of cities. 

%% CITY TYPOLOGY %%%
\subsubsection{City typology} \label{sec:city typology MIT}
As ground truth (typology) label for each city, we leverage the transportation related city typology labels provided by \cite{okeMIT} for 331 cities (worldwide). In the highest level of their hierarchical typologization, \cite{okeMIT} have six city typologies. For example, as shown in Table~\ref{tab:mit_typologies}, typologies such as auto or mass transit mainly denote high usage of respective modes in cities, whereas congestion typology indicate high level of congestion in the city. Each of the $331$ cities is assigned one of the six labels. Based on these typologies, we define four distinct city categories in our study \emph{i.e.}, `auto-heavy' (auto-dependent), `transit-heavy' (high usage of public transit), `bike-friendly' (high share of bike usage) and 'congestion'. For simplicity, we combine the mass transit and bus transit labels to form the `transit-heavy' label (\emph{i.e.}, a city is considered transit-heavy in our study if it is either labeled mass transit or bus transit in \cite{okeMIT}), and discard cities with the `hybrid' label in \cite{okeMIT}. This leaves us with $282$ cities, with each city having one of the four possible labels: congestion, auto-heavy, transit-heavy, and bike-friendly (metrobike). We adopt a one-versus-all approach where we focus on separately classifying a city for auto-heavy, transit-heavy, bike-friendly, and congestion labels (as explained in Section~\ref{sec:objective} below).

%%% OBJECTIVE %%%
\subsection{Objective} \label{sec:objective}
The objective of our study is to automatically answer the following questions about a city given its Wikipedia page.
\begin{enumerate}
    \item \text{Congestion prediction:} is the city congested?
    \item \text{Auto-heavy prediction:} are automobiles the major mode of transport for this city?
    \item \text{Transit-heavy prediction:} is public transit the major mode of transport for this city?
    \item \text{Bike-friendly prediction:} is bike a common form of transportation in this city?
\end{enumerate}

We focus on four separate classifiers; one for each of the tasks (questions) enumerated above. In other words, we adopt a one-versus-all approach to solve the problem of multi-class classification of a city (as assumed in \citealp{okeMIT}). Having separate classifiers brings in the flexibility to easily train, fine-tune, measure, and interpret the classifier specific to each typology type, and estimate the varying degree to which Wikipedia provides predictive information for each typology type of interest.
Moreover, separate scores from each of the classifiers are useful in cases where multiple labels for a city are allowed (\emph{e.g.}, it may be congested and also have public transit).

With the one-versus-all approach, we formulate four binary classification problems leading to the following conditional probability estimates for a city $i$ as \cref{eq:prob_c,eq:prob_a,eq:prob_t,eq:prob_b}.
\begin{align} %\label{eq:prob_estimates_formulation}
     \hat{p}^{(c)}_i =  & \mathbb{P}\left( \text{ city is congested } | \text{ city's Wikipedia page }  \right) \label{eq:prob_c} \\
    \hat{p}^{(a)}_i  = & \mathbb{P}\left( \text{ city is auto-heavy } | \text{ city's Wikipedia page }  \right) \label{eq:prob_a} \\
    \hat{p}^{(t)}_i = &\mathbb{P}\left( \text{ city is transit-heavy } | \text{ city's Wikipedia page }  \right) \label{eq:prob_t} \\
    \hat{p}^{(b)}_i = &\mathbb{P}\left( \text{ city is bike-friendly } | \text{ city's Wikipedia page }  \right) \label{eq:prob_b} 
\end{align}

The training objective for each of the four binary classifiers is the minimization of the binary-cross-entropy loss (log-loss) across all training samples (\citealp{kevin_murphy_ml_book}). 
Hence, the objective for the congestion prediction classifier can be stated as \cref{eq:cross_entropy}.
\begin{align} \label{eq:cross_entropy}
    \min \;\; - \sum_{i=1}^{n}  \left( (label^{(c)}_i) \ln{\left(\hat{p}^{(c)}_i \right)}  + (1-label^{(c)}_i) \ln{\left(1 - \hat{p}^{(c)}_i \right)}   \right)
\end{align}
where the sum is over all the training samples (of size $n$) , $label_i^{(c)} \in \{0,1\}$ is the congestion (binary) label for a city $i$ (\emph{i.e.}, the label is $1$ if city is congested, $0$ otherwise), and $\hat{p}^{(c)}_i$ is the probability estimate for the city being congested given the information in its Wikipedia page.
The objectives for the auto-heavy, transit-heavy, and bike-friendly classifiers can be written in a similar manner.

%%%% METHODOLOGY %%%
\section{Methodology} \label{sec:methodology}
In this section, we describe our proposed method in the context of the four classification tasks explained in Section~\ref{sec:objective}; the training data will be different for each of the tasks (based on the desired outcome), but the proposed supervised training method is the same for all the four tasks. We first give a high level overview of our proposed method in Section~\ref{sec:high level overview}.
This is followed by the details of supervised learning using logistic regression (Section~\ref{sec:LR}), keyline-based feature engineering (Section~\ref{sec:keyline feature engineering}), and infobox feature selection (Section~\ref{sec:infobox feature}).

%% HIGH LEVEL OVERVIEW %%
\subsection{High-level overview} \label{sec:high level overview}
In our supervised learning approach (for the congestion, auto-heavy, transit-heavy, and bike-friendly prediction problems), we first represent a city $i$ via a $5$-dimensional feature vector. The feature vector includes congestion, auto-heavy, transit-heavy, and bike-friendly keyline features and a numeric feature (\emph{i.e.}, population density). Intuitively, the congestion keyline feature indicates the presence of a line (text) in the city's Wikipedia page (main body) which strongly indicates that the city is congested; the term keyline is used since each line in the city's Wikipedia page is checked for semantic similarity with a pre-determined set of representative (key)lines indicating congestion.
The keyline features for auto-heavy, transit-heavy, and bike-friendly are designed in the same spirit, and the details are provided in Section~\ref{sec:keyline feature engineering}.
Eq. \ref{eq:city_features} represents the 5-dimensional feature vector $\mathbf{f}_i$ for a city $i$.
\begin{align} \label{eq:city_features}
    \mathbf{f}_i = 
    \begin{bmatrix}
    f^{(congestion)}_{i} \\
    f^{(auto)}_{i} \\
    f^{(transit)}_{i} \\
    f^{(bike)}_{i} \\
    f^{(density)}_{i} \\
    \end{bmatrix}
    {} &= \begin{bmatrix}
    f^{(c)}_{i} \\
    f^{(a)}_{i} \\
    f^{(t)}_{i} \\
    f^{(b)}_{i} \\
    f^{(density)}_{i} \\
    \end{bmatrix}
\end{align}
where $\mathbf{f}_i \in \mathbb{R}^5$, $f^{(c)}_i \in [-1, 1]$ denotes the congestion keyline feature, $f^{(a)}_i \in [-1, 1]$ denotes the auto keyline feature, $f^{(t)}_i \in [-1, 1]$ denotes the transit keyline feature, and $f^{(b)}_i \in [-1, 1]$ denotes the bike keyline feature (extracted from unstructured text in Wikipedia main body). Finally, $f^{(density)}_{i} \in \mathbb{R}$ denotes the population density of the city as inferred by the area and population details in the infobox section (structured text).

\begin{figure}[!htb]
\begin{center}
\includegraphics[scale=.50]{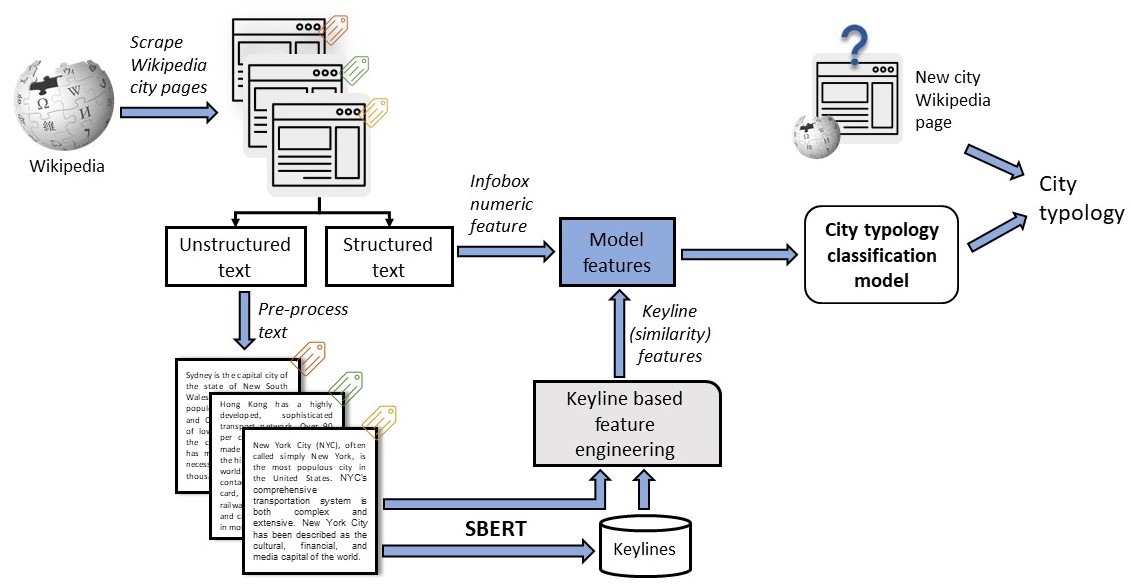}
\caption{High-level overview of our proposed method: steps include web scraping city Wikipedia pages, extracting numeric feature from structured text in the infobox and keyline similarity features from unstructured text in the main body (via keyline-based feature engineering), and training the typology classification model with these features to be used for predicting the city typology} \label{fig:highlevel_overview} 
\end{center}
\end{figure}

Given the above city representation, our approach is to train four logistic regression models to come up with the estimates $\hat{p}^{(c)}_i$, $\hat{p}^{(a)}_i$, $\hat{p}^{(t)}_i$, and $\hat{p}^{(b)}_i$ for a city $i$ as formulated in \cref{eq:prob_c,eq:prob_a,eq:prob_t,eq:prob_b}. An illustration of the above process is shown in Figure \ref{fig:highlevel_overview}.

Although we use logistic regression in this study (for simplicity and interpretability), the city features in \cref{eq:city_features} can be easily used with other linear models (\emph{e.g.}, support vector machine) and nonlinear models (\emph{e.g.}, decision trees, and neural networks) for classification.
A major methodological contribution in this study is the design of keyline features based on initial (typology representative) texts which we refer to as \textit{anchor texts} in our study (\emph{i.e.}, texts that are indicative of the intuitive meaning of the typologies; details in Table~\ref{tab:anchor_text}), and algorithmic keyline set expansion. This leads to a small set of representative lines (extracted from Wikipedia pages) which can be effective indicators of congestion, automobile usage, transit usage and bike usage in a city (details on keyline features and expansion method in Section~\ref{sec:keyline feature engineering}).

%% LOGISTIC REGRESSION %%
\subsection{Supervised learning using logistic regression}\label{sec:LR}
For brevity, we describe the logistic regression model in the context of estimating $\hat{p}^{(c)}_i$ for city $i$, \emph{i.e.}, the chances of city $i$ being congested given its Wikipedia page (the models for estimating $\hat{p}^{(a)}$, $\hat{p}^{(t)}$, and $\hat{p}^{(b)}_i$ can be described in a similar manner).
For estimating $\hat{p}^{(c)}_i$ via logistic regression, we consider the following parametric form in \cref{eq:sigmoid}.
\begin{align} \label{eq:sigmoid}
 \hat{p}^{(c)}_i  = \frac{1}{1 + e^{- ( \mathbf{w}^T \mathbf{f}_i + b) } } 
\end{align}
where the learning parameters $\mathbf{w} \in \mathbb{R}^5$, and bias $b \in \mathbb{R}$ are optimized by minimizing the binary cross entropy loss as defined in \cref{eq:cross_entropy}.
Note that for estimating $\hat{p}^{(a)}_i$, $\hat{p}^{(t)}_i$, and $\hat{p}^{(b)}_i$, we employ the same set of features $\mathbf{f}_i$ (as used for $\hat{p}^{(c)}_i$ above). In the following subsection, we describe the keyline features that occupy the first $4$ dimensions of $\mathbf{f}_i$.

%% KEYLINE FEATURE ENGINEERING %%
\subsection{Keyline-based feature engineering} \label{sec:keyline feature engineering}
To explain our proposed keyline-based features, we first explain the notion of semantic text similarity in Section~\ref{sec:semantic textual similarity}. Next in Section~\ref{sec:keyline_feature_computation}, we explain how starting from a manually chosen initial keyline (anchor text), we obtain a set of keylines extracted from Wikipedia pages for each of the four keyline features $f_i^{(c)}$,$f_i^{(a)}$,$f_i^{(t)}$, and $f_i^{(b)}$ mentioned in \cref{eq:city_features}.\\

%% semantic textual similarity %%
\subsubsection{Semantic textual similarity} \label{sec:semantic textual similarity}
The notion of semantic textual similarity (\citealp{sbert}) originated in the field of NLP, where the underlying task was to automatically identify if two sentences have similar meaning (\emph{i.e.}, one sentence is a paraphrased version of the other). For example, a perfect model for semantic textual similarity is expected to identify the following similar and dissimilar sentences:
\begin{itemize}
    \item Example 1: (this city is congested, this city suffers from traffic jams) $\rightarrow$ similar (score of $1$),
    \item Example 2: (this city is congested, this city does not experience traffic jams) $\rightarrow$ dissimilar (score of $-1$)
\end{itemize}

Note that semantic textual similarity is a very challenging task; even in the above example, a model needs to be intelligent enough to understand that \textit{congestion} and \textit{traffic jams} are related, while \textit{does not experience traffic jams} means there is no \textit{congestion}. As described in Sections~\ref{sec:introduction}\&\ref{sec:related work}, due to the introduction of SBERT, the state-of-the-art performance for semantic textual similarity tasks has seen a step-jump (thereby encouraging downstream applications like the one we propose in this study).
In our setup, we directly use the SBERT model fine-tuned for the semantic-textual-similarity task.  

SBERT can be used in the following manner  to estimate the semantic similarity between a pair of sentences (lines) $l_i$ and $l_j$ in \cref{eq:sim}.
\begin{align} \label{eq:sim}
    similarity(l_i, l_j) &= cosine\; similarity(\phi(l_i), \phi(l_j)) \nonumber\\
    &= \dfrac{\phi(l_i). \phi(l_j)}{norm(\phi(l_i))\times  norm(\phi(l_j))}\nonumber\\
    &= \widetilde{\phi(l_i)}^T\times \widetilde{\phi(l_j)}
\end{align}
where $\phi(l) \in \mathbb{R}^{768}$ denotes the $768$-dimensional sentence-embedding for sentence $l$ using the trained sentence-BERT model in \cite{sbert}. Eq. \ref{eq:sim} is essentially the cosine-similarity between the embeddings of the two sentences (and is in the range $[-1,1]$). The cosine similarity between two vectors can be defined as the inner product of the same vectors normalized to both have length 1. The sentence embedding $\phi(l)$ is normalized to have unit $\ell_2$ norm; $\widetilde{\phi(l)}$ denote the normalized vector and $T$ denotes the transpose operation. Essentially, the cosine similarity between two lines ($l_i$ and $l_j$) can be calculated using matrix multiplication on the two normalized vectors ($\widetilde{\phi(l_i)}^T$ and $\widetilde{\phi(l_j)}$). Based on this calculation, two sentences with similar meaning will have higher scores than two non-similar sentences. 
To give a quantitative feel, Table~\ref{tab:sbert_examples} provides a few examples of pairs of sentences and their SBERT similarity scores (obtained using \cref{eq:sim}). Just for the purposes of an illustrative comparison, for the same pair of sentences, we obtain a Word2Vec based sentence representation (\emph{i.e.}, the average of Word2Vec embeddings of constituent words to represent a sentence), and compute the similarity score. In other words, for a sentence $l$ in Table~\ref{tab:sbert_examples}, we compute $\phi(l)$ using SBERT and Word2Vec embedding models and use these vectors to calculate the corresponding pair-wise sentence similarity scores (using \cref{eq:sim}).
As shown in Table~\ref{tab:sbert_examples}, SBERT performs much better compared to Word2Vec in gauging the semantic similarity between sentences, and this observation around the superiority of SBERT compared to prior sentence representation methods (\emph{e.g.}, Word2Vec, Doc2Vec, TF-IDF, GloVe, FastText) is consistent with findings in the wider NLP literature (\citealp{sbert,bert}). 
   \begin{table}[h]
    \centering
    \caption{Illustrative examples for sentence-BERT and Word2Vec based similarity scores for pairs of sentences; the sentence-BERT scores are calculated using the %SentenceTransformers model version
    stsb-distilbert-base version in (\cite{sbert}), while the Word2Vec scores are computed using the Spacy package \cite{spacy} in Python.}
    \begin{tabular}{|l|c|c|c|}
    \hline
     Line 1   & Line 2 & SBERT score & Word2Vec score \\
     \hline
        I am a student      &  I am enrolled in a university  & 0.767 & 0.918 \\
        I am a student      &  I am not enrolled in a university  & 0.356  & 0.917 \\
        The city is car dependent & A lot of people in the city use automobile & 0.663 & 0.877 \\
         The city is car dependent & Many people do not use automobile in the city & 0.280 &  0.847 \\
    \hline
    \end{tabular}
    \label{tab:sbert_examples}
\end{table}

\textcolor{black}{However, availability of city information on Wikipedia and existing NLP techniques like SBERT does not itself solve the problem of large-scale city typology classification. One of the key challenges in using such deep learning models is the limited data availability (e.g., lack of open data on city related features and limited ground truth labels on city typologies). This is because training such models for a specific task requires thousands of input samples and needs proper feature engineering. Therefore, the keyline-based feature engineering component of the proposed methodology addresses the problem of efficiently extracting relevant information from Wikipedia and effectively using this data (combining both textual and numeric information) to develop prediction models with limited ground truth labels.} We explain below, how such pair-wise semantic similarity measures obtained via SBERT can be used to derive city level features from Wikipedia pages.\\

\subsubsection{Keyline similarity features} \label{sec:keyline_feature_computation} 
\begin{figure}[!htb]
\begin{center}
\includegraphics[scale=.49]{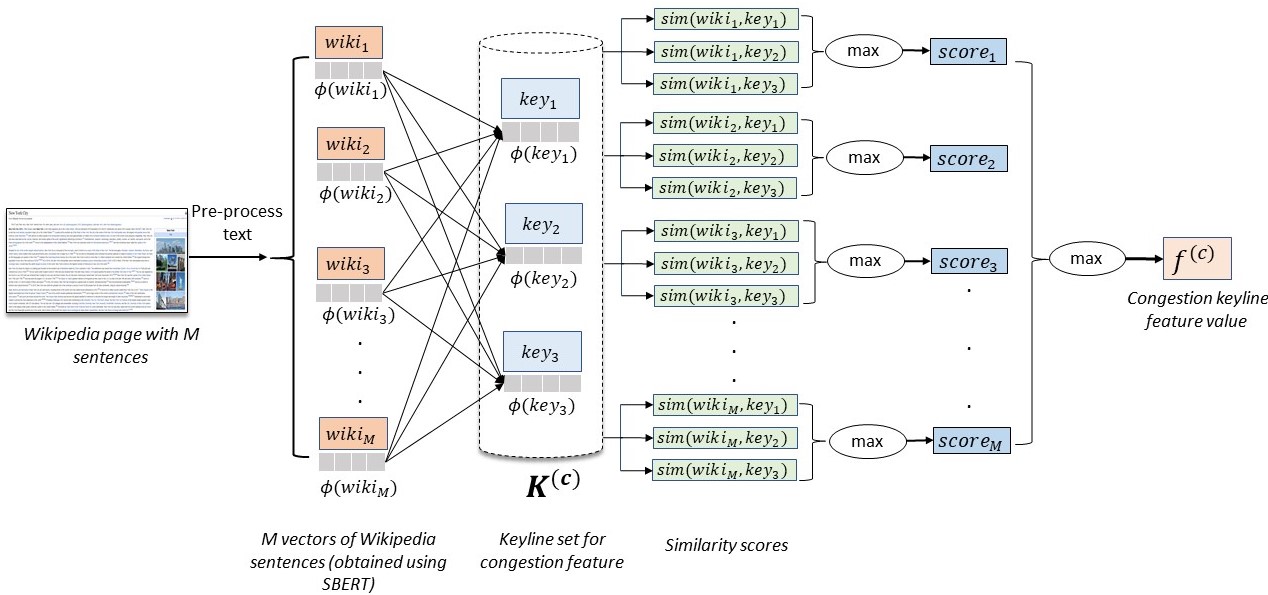}
\caption{Illustration of congestion keyline similarity feature extraction from a city's Wikipedia page using the congestion keyline set, assuming that the set has $3$ representative keylines indicating congestion.} \label{fig:keyline_similarity_features}
\end{center}
\end{figure}

Assuming the presence of a semantic-textual-similarity model as described above in Section~\ref{sec:semantic textual similarity}, we focus on the following idea. Consider the congestion prediction task (defined in Section~\ref{sec:objective}), where one is given the Wikipedia page of a city, and has to estimate the chances of the city being congested.
Intuitively, if the Wikipedia page has line(s) which are semantically similar to \textit{the city suffers from traffic congestion}, there may be a good chance that the city is indeed congested.
However, there may be other ways in which the city's congestion problem may be expressed in the Wikipedia page. For example, \textit{cars are stuck for hours on the main roads of the city on weekdays} is another plausible (key)line representing congestion.
Building on this intuition, if we can construct a
small yet representative set of keylines indicating congestion in a city, we can check each line in a city's page (as shown in the bi-partite graph in Figure~\ref{fig:keyline_similarity_features}) with the representative keylines to see if there is a high semantic similarity; having a set of keylines just casts a wider net compared to having just one keyline. The highest semantic similarity score across all possible pairs is derived from the keylines and the city's Wikipedia page lines can then serve as a \textit{congestion} keyline feature for the city (\emph{i.e,}, $f_i^{(c)}$ for a city $i$), as illustrated in Figure~\ref{fig:keyline_similarity_features}. This is precisely the idea behind the congestion keyline feature proposed in this study and we give a formal description below.

Consider the Wikipedia page of city $i$, and the indexed set of lines $\{wiki_{i1}, wiki_{i2} , \ldots , wiki_{iM_i} \}$ in the main body of the page (assuming the page $i$ has $M_i$ lines of text).
Using \cref{eq:sim} to compute semantic similarity between two sentences, the congestion keyline feature can be computed as \cref{eq:congestion-keyline-feature}.
\begin{align} 
    f^{(c)}_i = \max_{ 1 \leq j \leq M_i } \max_{ 1 \leq k \leq |\mathbf{K}^{(c)}|  } similarity \left (wiki_{ij}, key^{(c)}_k\right)  \label{eq:congestion-keyline-feature}
\end{align}
where $wiki_{ij}$ is the $j$-th line in the city $i$'s Wikipedia page, $key^{(c)}_k$ is the $k$-th keyline in the set of keylines for congestion (set denoted by $\mathbf{K}^{(c)}$), and the feature value $f^{(c)}_i$ is the maximum cosine similarity between all Wikipedia page lines of the city, and all the keylines in the congestion keyline set $\mathbf{K}^{(c)}$ (similarity calculated using \cref{eq:sim}).\\
The auto, transit, and bike keyline features can be stated in a similar manner as shown in \cref{eq:auto-keyline-feature,eq:transit-keyline-feature,eq:bike-keyline-feature}.
\begin{align}
    f^{(a)}_i = \max_{ 1 \leq j \leq M_i } \max_{ 1 \leq k \leq |\mathbf{K}^{(a)}|  } similarity \left (wiki_{ij}, key^{(a)}_k   \right) \label{eq:auto-keyline-feature} ,  \\
    f^{(t)}_i = \max_{ 1 \leq j \leq M_i } \max_{ 1 \leq k \leq |\mathbf{K}^{(t)}|  } similarity \left  (wiki_{ij} , key^{(t)}_k \right) \label{eq:transit-keyline-feature} ,  \\
    f^{(b)}_i = \max_{ 1 \leq j \leq M_i } \max_{ 1 \leq k \leq |\mathbf{K}^{(b)}|  } similarity \left (wiki_{ij} , key^{(b)}_k  \right ) \label{eq:bike-keyline-feature}, 
\end{align}
where $\mathbf{K}^{(a)}$, $\mathbf{K}^{(t)}$, and $\mathbf{K}^{(b)}$ denote the auto, transit and bike keyline sets respectively. For computational efficiency, the above features can be computed via a matrix multiplication (between a matrix of stacked embeddings of the Wikipedia page lines and matrix of stacked keyline embeddings).

The above feature computations assume a set of keylines for congestion, auto, transit and bike. We describe below an algorithm to obtain such keyline sets starting from an initial guess and iteratively extracting lines from Wikipedia pages of cities in the training dataset.\\

%% KEYLINES SET (CANDIDATES AND EXPANSION) %%
\subsubsection{Keyline sets (initial guess and set expansion)} \label{sec:method_expansion}
We describe below our proposed method for constructing the set $\mathbf{K}^{(c)}$, \emph{i.e.}, keylines for congestion (the method for $\mathbf{K}^{(a)}$, $\mathbf{K}^{(t)}$, $\mathbf{K}^{(b)}$ is similar and their description has been skipped for brevity). To construct the keyline sets $\mathbf{K}^{(c)}$, $\mathbf{K}^{(a)}$, $\mathbf{K}^{(t)}$, $\mathbf{K}^{(b)}$ we start with initial guesses for these sets (one line for each set).
These initial keyline guesses will be referred to as anchor text, and our choices in this study are listed in Table~\ref{tab:anchor_text}.

\begin{table}[h]
    \centering
    \caption{Initial keylines (anchor text) for the city typology prediction tasks considered in this study}
    \begin{tabular}{|l|l|l|}
    \hline
     Keyline feature type &  Initial keyline or anchor text & Notation \\
     \hline
        congestion       & `the city has heavy traffic congestion'    &  ${anchor}^{(c)}$  \\
        auto     &   `most people in the city use cars'   &  ${anchor}^{(a)}$  \\
        transit   &  `most people in the city use public transit like bus and metro'   &  ${anchor}^{(t)}$ \\
        bike & `many people in the city use bike or cycle'  &  ${anchor}^{(b)}$\\
    \hline
    \end{tabular}
    \label{tab:anchor_text}
\end{table}

The choice of words in the anchor text may appear arbitrary (more like an initial guess) but they all allude to the underlying meaning of the feature. For example, the congestion anchor text intends to capture Wikipedia page lines which may indicate that the city is congested, while the transit anchor text intends to capture the dominance of public transit services in the city. It is plausible that these choices are not representative enough, and to cast a wider net we propose a keyline set expansion method which automatically extracts representative lines from the Wikipedia pages in the training data and expands the keyline set (thereby reducing the dependency on the initial guesses to a large extent).

The keyline set expansion algorithm assumes that we have a dataset consisting of Wikipedia cities, and their congestion labels (\emph{i.e.}, congested or not congested). We randomly divide the dataset into train and test sets, and use only the train data to extract additional keylines.
Specifically, we focus on all cities in the train set which are labeled positive (\emph{i.e.}, congested). From each positively labeled city $i$ in the train set, we extract a \textit{candidate} congestion keyline as shown in \cref{eq:candidate_keyline_c}.

\begin{align} \label{eq:candidate_keyline_c} 
    candidate\;keyline^{(c)}_i = \text{arg} \max_{ 1 \leq j \leq M_i }  similarity \left(wiki_{ij} , anchor^{(c)}    \right )  
\end{align}
where $anchor^{(c)}$ is the anchor text for the congestion feature as listed in Table~\ref{tab:anchor_text}. $candidate\; keyline^{(c)}_i$ is the semantically closest line in the page of city $i$ compared to the anchor text for congestion. For example, a candidate keyline obtained from the Wikipedia page of the city Manila is \textit{Manila is notorious for its frequent traffic jams and high densities.}

 Using similar notations as in \cref{eq:candidate_keyline_c}, candidate keylines for auto, transit, and bike are extracted using \cref{eq:candidate_keyline_a}, \cref{eq:candidate_keyline_t}, and \cref{eq:candidate_keyline_b} respectively.
\begin{align} 
    candidate\;keyline^{(a)}_i = \text{arg} \max_{ 1 \leq j \leq M_i }  similarity \left(wiki_{ij} , anchor^{(a)}    \right )\label{eq:candidate_keyline_a}  \\
     candidate\;keyline^{(t)}_i = \text{arg} \max_{ 1 \leq j \leq M_i }  similarity \left(wiki_{ij} , anchor^{(t)}    \right )\label{eq:candidate_keyline_t} \\
      candidate\;keyline^{(b)}_i = \text{arg} \max_{ 1 \leq j \leq M_i }  similarity \left(wiki_{ij} , anchor^{(b)}    \right )\label{eq:candidate_keyline_b} 
\end{align}

\begin{figure}[!htb]
\begin{center}
\includegraphics[scale=.51]{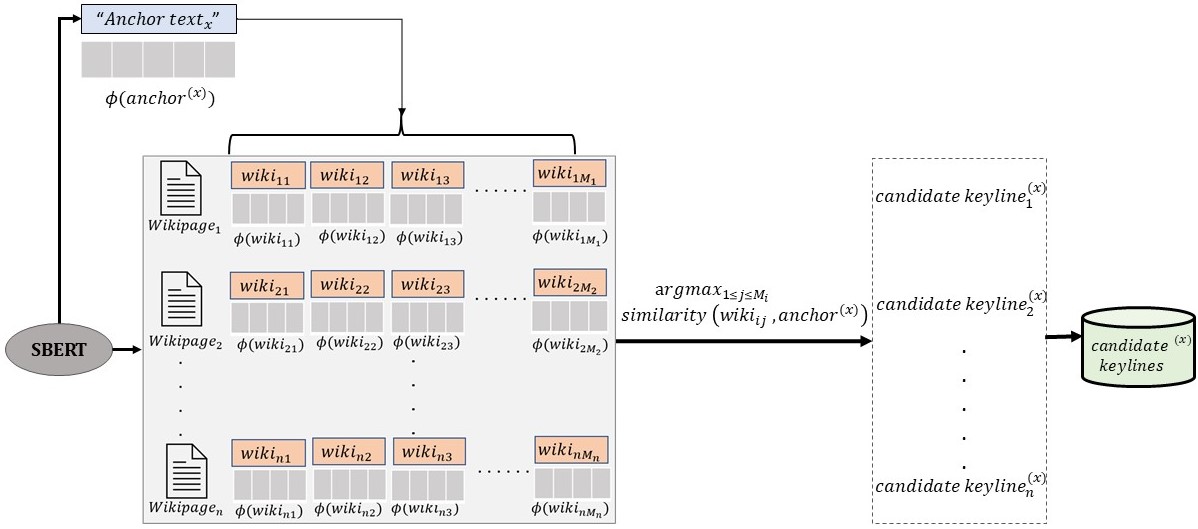}
\caption{Selection of candidate keylines for a feature type $x$ from $n$ Wikipedia pages. $x$ can be congestion, auto, transit or bike feature. For a city $i$ with $M_i$ sentences in its Wikipedia page, each sentence $l$ is converted to 768-dimension vector ($\phi(l)$) and so is the feature $x$ anchor text (using sentence-embedding by SBERT)} \label{fig:candidate_keylines}
\end{center}
\end{figure}

For each feature, we collect all such candidate keylines from the training dataset (as shown in Figure~\ref{fig:candidate_keylines}), and sort them in decreasing order of similarity score with their anchor texts. For congestion we obtain the list of sorted candidate keylines as $sorted\_candidates^{(c)}$.

The keyline set expansion method for the feature \textit{congestion} is outlined in Algorithm 1. We first initialize the feature keyline sets with the respective anchor texts (singleton keyline sets). We then go over all candidates in the ordered $sorted\_candidates^{(c)}$ list (in the train set), and greedily add one candidate at a time to the keyline set $\mathbf{K}^{(c)}$. In each iteration $e$ (where we add a candidate), we derive the keyline similarity feature $f^{(c)}$ from the current iteration's keyline set $\mathbf{K}^{(c)}$ (using \cref{eq:congestion-keyline-feature}). The $f^{(a)}$, $f^{(t)}$, and $f^{(b)}$ features for training the logistic regression model in each iteration are derived from only the anchor texts in the corresponding singleton keyline sets $\mathbf{K}^{(a)}$, $\mathbf{K}^{(t)}$, and $\mathbf{K}^{(b)}$ (using \cref{eq:auto-keyline-feature,eq:transit-keyline-feature,eq:bike-keyline-feature}). Using the computed features, we train the logistic regression model for the congestion prediction task (as explained in Section~\ref{sec:LR}).
Using the trained logistic regression model on the validation set, we record the validation set performance metric (AUC, \emph{i.e.}, area under the receiver operating characteristic curve for binary classification predictions, details in Section~\ref{sec:metrics}). We keep track of the validation metric across all iterations, and finally select the iteration (and the corresponding keyline set)
with the best validation set performance to arrive at the optimized (and expanded) keyline set for congestion. %choose the set which is maximum across three folds

%{\SetAlgoNoLine%
\begin{algorithm}[H] \label{alg:expansion}
\textbf{Input:} train set, validation set, $\mathbf{K}^{(a)}$ = \{$anchor^{(a)}$\}, $\mathbf{K}^{(t)}$ = \{$anchor^{(t)}$\}, $\mathbf{K}^{(b)}$ = \{$anchor^{(b)}$\}, 
$num\_candidates$ = $\|$$sorted\_candidates^{(c)}$$\|$ in train set\\
\textbf{Output:} max\_expansion, optimized expanded keyline set $\mathbf{K}^{(c)}_{opt}$

\begin{algorithmic}[1]
%\begin{ALC@g}
\STATE \textbf{Initialization}: max\_AUC = 0, candidate\_index = 1, max\_expansion = 0, $\mathbf{K}^{(c)}$ = \{$anchor^{(c)}$\}\\
\STATE \While{candidate\_index $\leq$ num\_candidates}
   {
    \Indp \STATE ADD candidate from $sorted\_candidates^{(c)}$ to congestion keyline set $\mathbf{K}^{(c)}$,\\ 
    \STATE COMPUTE keyline similarity features: $f^{(c)}$, $f^{(a)}, f^{(t)}$, and $f^{(b)}$ from $\mathbf{K}^{(c)}$, $\mathbf{K}^{(a)}$, $\mathbf{K}^{(t)}$, and $\mathbf{K}^{(b)}$  respectively (using \cref{eq:congestion-keyline-feature,eq:auto-keyline-feature,eq:transit-keyline-feature,eq:bike-keyline-feature})
    \STATE TRAIN congestion classification model using train set, compute %train set AUC and 
    validation AUC (using validation set)
    \STATE \If{validation AUC $>$ max\_AUC}{
        max\_AUC   $\gets$  validation AUC, \;     max\_expansion $\gets$ candidate\_index}
    \STATE \Else{
        \textbf{continue}}
    \STATE candidate\_index $\gets$ candidate\_index + 1 
    }
\STATE $\mathbf{K}^{(c)}_{opt}$ = $\mathbf{K}^{(c)}$[: max\_expansion+1] %$\mathbf{K}^{(x)}_{initial}$ $\bigcup$ 
%\end{ALC@g}
\end{algorithmic}

\caption{Greedy keyline set expansion for congestion keyline feature (using validation set performance)}
\end{algorithm}%}%

Note that the (typology) classification model used in the keyline expansion method is based on the feature whose keyline set is to be expanded. For example, if the method is applied to the congestion feature in order to get a representative set of keylines indicating congestion ($\mathbf{K}^{(c)}$), then the congestion typology classification model is used (as shown in Algorithm 1). Similarly, we use auto-heavy typology classifier for computing $\mathbf{K}^{(a)}$, transit-heavy typology classifier for $\mathbf{K}^{(t)}$, and bike-friendly typology classifier for $\mathbf{K}^{(b)}$. 

To assess performance of the LR models used in the keyline set expansion method (Algorithm 1 step 5), we perform 3-fold cross-validation. Cross-validation gives an idea about how well the trained model will generalize for an unseen dataset, and avoids fitting the model to just the training dataset.
To do this, at each iteration $e$, we split our training dataset into three equal parts; for each instance (part) in our dataset, we build a logistic regression model using all other instances and then validate it on the selected instance (\emph{i.e.,} validation set). In our setup, the 3-fold cross validation process is repeated three times for each iteration, and the mean AUC value across all folds from all runs
is considered as the validation AUC at that iteration (step 5 in Algorithm 1). This is done to reduce error in the estimate of mean performance of the model. 

Note that the test data set is kept aside for final evaluation, and only the validation data set is used to optimize keyline set expansion. The keyline set expansion algorithm for auto, transit, and bike keyline features is similar, and we skip their description for brevity. Also, one may suggest considering all the candidate keylines from the training set (say, $\mathbf{K}^{(c)}_{all}$ for congestion feature) instead of the optimized feature keyline set ($\mathbf{K}^{(c)}_{opt}$ obtained using Algorithm 1 for congestion). However, using $\mathbf{K}^{(c)}_{all}$ for feature definition is not only computationally costly but also results in sub-optimal performance on the test set compared to the optimal keyline set (details in Section~\ref{sec:results}). 

%% NUMERIC INFOBOX DATA %%
\subsection{Infobox numeric feature} \label{sec:infobox feature}
To obtain the population density feature ($f_i^{(density)}$) for a city $i$, the density details are obtained from the infobox in the city's Wikipedia page.  For pages with no density information, we simply extract the recent population estimate and area of the city from the infobox, and compute the city's population density (population/area) to be used as a numeric feature 
in the typology classification models along with other keyline features.

\section{Experiments}\label{sec:experiments}
In this section, we provide details around the data (Section~\ref{sec:data}) and evaluation metrics  used in our experiments (Section~\ref{sec:metrics}). This is followed by a description of the initial keyline features (Section~\ref{sec:keyline_expansion_cv}) computed using the anchor texts.

\subsection{Data} \label{sec:data} 
In our study we focus on four binary classification tasks centred around: congestion, auto-heavy, transit-heavy, and bike-friendly typology prediction for a city.
For training each of the above binary classifiers in a supervised manner, we need ground truth labels in the form of ($city_i$, $label^{(task)}_i$) where binary $label^{(task)}_i \in \{ 0,1\}$, and $task$ specifies the classification task (congestion, auto-heavy, transit-heavy, and bike-friendly).
We obtain the task specific binary labels (ground truth) for a city following a recent study \cite{okeMIT} as described below.

As described in Section~\ref{sec:city typology MIT}, we obtain $282$ cities with each city having one of the four possible labels: congestion, auto-heavy, transit-heavy, and bike-friendly. We will refer to this dataset as the 'typology' dataset. The typology label distribution across these $282$ cities is as follows: 27\% congestion, 23\% auto-heavy, 39\% transit-heavy, and 11\% bike-friendly. Using these four labels, we compute the task-specific binary labels for a city in a one-versus-all fashion, \emph{e.g.}, for the auto-heavy prediction task, all cities with the auto-heavy label are assigned a label of $1$, and the remaining cities (from the remaining $3$ typologies) are assigned a label of $0$.

Once the city typology dataset is finalized as described above, we collect their Wikipedia addresses (URLs) using the \cite{wikiapi}. Using the city URLs, we automatically crawl the Wikipedia pages associated with the URLs, and collect data from the main bodies and infoboxes. We use web scraping tools in Python (such as Beautiful Soup, Wikipedia API, and Pandas) for data cleaning and processing as described below.\\

\begin{figure}[!htb]
\begin{center}
\includegraphics[scale=.49]{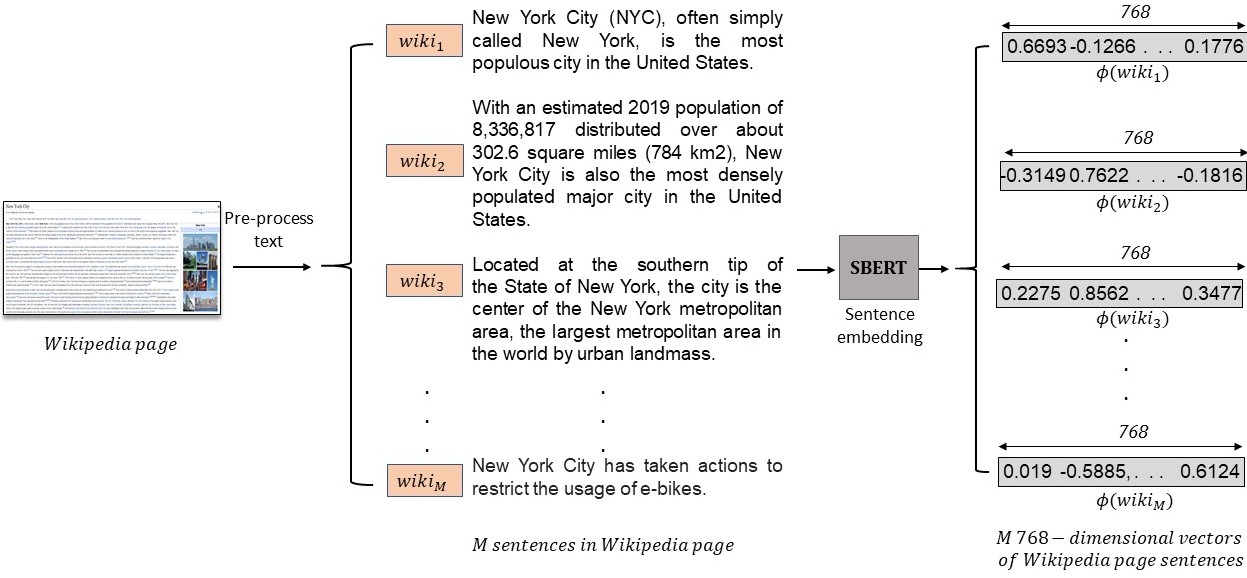}
\caption{Textual data extraction from New York City Wikipedia page main body with $M$ sentences and vector representation of these sentences using pre-trained SBERT model.} \label{fig:sbert_vec_embedding}
\end{center}
\end{figure}

\paragraph*{Data from unstructured main body} The qualitative information on each city in the typology dataset are collected from different sections in the respective Wikipedia pages (including sections like demography, geography, economy, transportation, infrastructure, education). Although we focus on the transportation aspect of cities, we chose to collect data from all sections in a Wikipedia page, and not just the transportation section. This is because, in some cases, information regarding mobility scenario in a city is present in multiple sections such as infrastructure, and economy. The textual data extracted for each city are pre-processed (such as removing section titles, footnotes and references) and stored in a format such that given a city name, we can get a list of sentences from the city's Wikipedia page. Each sentence $l$ in this list is converted to a $768$-dimensional real-valued vector ($\phi_l \in \mathbb{R}^{768}$) using a pre-trained SBERT model (using the version stsb-distilbert-base in \cite{sbert} pre-trained for semantic textual similarity). In this manner, for each city $i$, we obtain a list of $M_i$ vectors (with each vector being $768$-dimensional); 
here $M_i$ denotes the number of sentences extracted from the Wikipedia page of the city $i$. Figure~\ref{fig:sbert_vec_embedding} provides an illustration of this process for New York City. \\

\paragraph*{Data from structured infobox} The quantitative information pertaining to each city's demography are collected from the corresponding Wikipedia infobox. In particular, we extract population density, population and area estimates (all in text format), pre-process the data, and convert them to numeric values. The density and area values in Wikipedia articles are available in both mile and kilometer units. For cities with no density information in their Wikipedia pages, we compute the missing values using their population and area estimates. The derived population densities (in $sq.mi$) for each city are first normalized and then used as numeric features in the typology classification models in our study. Additionally, the city coordinates are also obtained from their respective Wikipedia pages for visualization purposes.\\

\paragraph*{Train and test datasets} 
For our typology prediction tasks, we use logistic regression (LR) model for each of the typology classifiers. Such models are trained (supervised) based on a training dataset so that the model learns the relationship between the input and output variables; the performance of the model is then assessed using a test dataset (different from training set yet representative of the dataset as a whole). Therefore, in our case we consider 70\%  of the city typology data for training (\emph{i.e.}, 197 cities)
%with 28\% congestion, 23\% auto-heavy, 36\% transit-heavy, and 13\% bike-friendly cities) 
and 30\%  for testing purposes (\emph{i.e.}, 85 cities; details summarized in Table~\ref{tab:data_distribution}. %with  23\% congestion, 23\% auto-heavy, 46\% transit-heavy, and 8\% bike-friendly cities). 
For both the train and test data sets, we generate positive and negative labels (1s and 0s) for each typology classification model (binary classifier) based on the respective typology. For example, for \textit{auto-heavy} model, we label cities with category $auto$ as 1 while cities with categories other than $auto$ (\emph{e.g.,} congestion, transit or bike) are labeled 0. Therefore, the train set (197 cities) and the test set (85 cities) used in each classification model remain the same. However, the city typology labels (output variables) are modified based on the prediction task of the LR models (\textcolor{black}{study data and related code are available in Zenodo, please see \citealp{zenodo})}.

%%%% Sample distribution table %%%%%%%%%%%%%%%%%
\begin{table}[h]
    \centering
    \caption{Experimental data distribution (for 282 cities considered in our study)}
    \begin{tabular}{c|c|c|c|c|c}
    \hline
    \multirow{2}{*}{Data} & \multirow{2}{*}{Sample size} & \multicolumn{4}{c}{Category distribution}\\
    \cline{3-6}
     &  & Congestion & Auto-heavy & Transit-heavy & Bike-friendly\\
     \hline
    Train set & 197 & 28\% & 23\% &  36\% & 13\%\\
    \hline
    Test set & 85 & 23\% & 23\% & 46\% & 8\%\\
    \hline
    \end{tabular}
    \label{tab:data_distribution}
\end{table}

%% METRICS %%
\subsection{Evaluation metrics}  \label{sec:metrics}
The idea behind testing supervised learning models on a new set of observations (\emph{i.e.}, test set) is to examine whether the trained model is able to generalize well to be able to effectively perform the task with new data.
For quantitatively measuring the performance of the typology classifiers in our study, we consider the metric widely used for evaluating binary classification models \emph{i.e.,} AUC (Area under the curve) ROC (Receiver Operating Characteristics) curve (\citealp{kevin_murphy_ml_book}). A binary classification model can be used to predict the probability of a data sample belonging to one class or the other. Based on such a predicted probability and a threshold, one can identify positive and negative labels for each of the data samples. The proportion of the positive class that gets correctly classified by the model is the true positive rate (TPR \emph{i.e.,} ratio of true positive to all positives), while the proportion of the negative class which the model incorrectly classified is the false positive rate (FPR \emph{i.e.,} ratio of false positives to all negatives). Therefore, different values of thresholds consequently affect the TPR and FPR. The ROC curve is a probability curve that plots the TPR against FPR at various threshold values. 
AUC provides the summary of the ROC curve and can be used to compare classifiers directly without specific decision thresholds. In simple terms, the AUC score (AUC $\in$ $[0,1]$) tells us how well the model is able to distinguish between positive and negative classes. For example, consider the congestion classification task: the higher the AUC, the better the model is at classifying between whether the city is congested or not. In our study, we consider AUC as the main evaluation metric to assess a model's performance (as described in sections below). 

In addition, for the models with features obtained from optimal expanded keyline sets, we report the accuracy (fraction of samples where predicted label matches ground truth label), precision (true positives over the sum of true positives and false positives), recall (true positives over the sum of true positives and false negatives), and F-1 score (harmonic mean of precision and recall) (\citealp{kevin_murphy_ml_book}) in Section~\ref{sec:performance}. Based on the probability scores obtained using an LR model, the classification scores (accuracy, precision, recall, and F1-score) are computed based on a decision (probability) threshold. This threshold governs the decision to convert a probability score into a class label (probability values higher than the threshold are assigned label 1 else 0 in binary classification). The optimal decision threshold can be obtained by doing a search over thresholds in $[0,1]$. The optimal threshold for each model is selected based on the corresponding ROC curve and the largest G-mean score, \emph{i.e.}, the geometric mean between TPR and true negative rate (1 - TPR); this approach is commonly used for imbalanced classification.

%% MODEL FEATURES %%
\subsection{Initial keyline features} \label{sec:keyline_expansion_cv}
For each city in our data we obtain a list of 768-dimension vectors (the size of the list varies based on the number of lines in each city Wikipedia page). 
Our proposed method of algorithmically extracting lines from a city’s Wikipedia page (Algorithm~1) to semantically match the typology of interest provides a $4$-dimensional keyline-based feature vector for each city (as mentioned in \cref{eq:city_features}). As outlined in Algorithm 1, the feature keyline sets ($\mathbf{K}^{(c)}, \mathbf{K}^{(a)}, \mathbf{K}^{(t)}$, and $\mathbf{K}^{(b)}$) are initialized using their corresponding anchor texts. For clarity, we denote these singleton keyline sets as $\mathbf{K}_{initial}^{(c)}$, $\mathbf{K}_{initial}^{(a)}$, $\mathbf{K}_{initial}^{(t)}$, and $\mathbf{K}_{initial}^{(b)}$. The associated keyline features ($f^{(c)}$, $f^{(a)}$, $f^{(t)}$, and $f^{(b)}$) for each city are computed using the keyline similarity formulae (\cref{eq:congestion-keyline-feature,eq:auto-keyline-feature,eq:transit-keyline-feature,eq:bike-keyline-feature}). We denote these features as $f_{initial}^{(c)}$, $f_{initial}^{(a)}$, $f_{initial}^{(t)}$, and $f_{initial}^{(b)}$;  values of these features for example cities in the typology data are shown in Table~\ref{tab:initial_model_features}.
\begin{table}[h]
    \centering
    \caption{Keyline feature values (obtained using the anchor texts) for example cities in the typology data}
    \begin{tabular}{|l|c|c|c|c|}
    \hline
    City name & $f_{initial}^{(c)}$ & $f_{initial}^{(a)}$ & $f_{initial}^{(t)}$ &  $f_{initial}^{(b)}$\\
    & & & &\\
     \hline
        Dhaka, Bangladesh	&	0.681	&	0.500	&	0.525	&	0.550\\
        Dubai, United Arab Emirates	&	0.475	&	0.429	&	0.504	&	0.441\\
        Amsterdam, Netherlands	&	0.533	&	0.493	&	0.585 &	0.776	\\
        Changchun, China	&	0.510	&	0.485	&	0.449	&	0.427\\
    \hline
    \end{tabular}
    \label{tab:initial_model_features}
\end{table}

The typology classification models trained using these (anchor text-based) keyline features serve as base models in our study. Note that since we have at most $282$ labeled samples for training a classifier, we cannot directly use a $768$-dimensional representation of the Wikipedia page by simply averaging the vectors across lines in the Wikipedia page. 
A common rule of thumb for training a logistic regression model is to have at least 10 times more data samples than the number of weights the model has to learn (feature dimension). In comparision, our proposed 4-dimensional feature vector makes the training tractable for the small size of our dataset.
In addition, our feature vector is easily interpretable (connected with typologies) as opposed to a generic dimensionality reduction technique like principal component analysis \cite{kevin_murphy_ml_book}.

%% RESULTS %%
\section{Results and Discussion} \label{sec:results}
In this section, we discuss the results from the generated feature candidate keylines (Section~\ref{sec:generated candidate keylines}), feature keyline set expansion (Section~\ref{sec:keyline set expansion results}) and typology classification models (Section~\ref{sec:performance}). This is followed by insights and applications in Section~\ref{sec:insights}.

\subsection{Generated candidate keylines}\label{sec:generated candidate keylines} Our proposed method considers using a set of representative keylines pertaining to a city feature (in addition to its anchor text). First, we obtain the candidate keylines related to each feature (as described in Section~\ref{sec:method_expansion}). Using the train data prepared for each typology classifier (as in Section~\ref{sec:data}) and the feature anchor texts ($anchor^{(c)}, anchor^{(a)}, anchor^{(t)}$ and $anchor^{(b)}$), we extract candidate keylines pertaining to congestion, auto, transit, and bike features using \cref{eq:candidate_keyline_c,eq:candidate_keyline_a,eq:candidate_keyline_t,eq:candidate_keyline_b} respectively. 

The number of candidate keylines obtained for congestion, auto, transit, and bike features  are 56, 45, 71, and 25 respectively; these numbers correspond to the typology distribution in the train set (\emph{e.g.}, for auto we have 45 positive samples with auto-heavy typology in the train set, and one candidate keyline is derived from the Wikipedia page corresponding to each of the positive samples). An example candidate auto keyline is \textit{many of these auto routes are frequently congested at rush hour}; this is obtained from Montreal (Canada) Wikipedia page (having maximum similarity with the auto anchor text \emph{i.e.}, \textit{most people in the city use cars}). The similarity scores of the candidate keylines (\emph{i.e.}, with anchor texts) range between 0.31 to 0.79.  Our experiments are carried out on a computer with Intel i$7$ processor with $2$ cores, $4$ logical processors and $16$ GB RAM. The computation time noticed for the above mentioned sets of candidate keylines is 22 minutes for congestion, 25 minutes for auto, 28 minutes for transit, and 13 minutes for bike. Therefore, average computation time for finding a keyline (indicative of a typology) on a Wikipedia page using our method is around 30 seconds. 
Candidate keylines are able to effectively capture relevant signals indicative of the typologies from the Wikipedia pages; some variability is noticed in keylines with lower scores which may be attributed to the amount and type of information present in corresponding Wikipedia pages. The effect of such noise in candidate keylines on the typology prediction tasks is reduced with the use of optimal keyline sets derived from these candidate keylines (using our proposed feature keyline set expansion method). 
%Having this set of representative keylines indicative of the typology casts a wider net for retrieving relevant signals from the city Wikipedia pages and reduces the dependency on a single keyline. 
We sort the candidate keylines (in descending order by similarity score) and store them as $sorted\_candidates^{(c)}$,
$sorted\_candidates^{(a)}$, $sorted\_candidates^{(t)}$, and
$sorted\_candidates^{(b)}$ respectively for feature keyline set expansion.

%Once the candidate keylines pertaining to congestion, auto, transit, and bike (number of candidate keylines $\approx$ 56, 45, 71, and 25 respectively) are obtained, 

%Congestion score range 0.37-0.73, Auto 0.31-0.67; Transit 0.36-0.76 ; Bike 0.31-0.79.

%% KEYLINE SET EXPANSION RESULTS %%
\subsection{Feature keyline set expansion results} \label{sec:keyline set expansion results}
The keyline set expansion method for a feature (as outlined in Algorithm~1) requires that at each expansion iteration ($e$), we expand the selected feature keyline set (step 3), compute associated feature using the expanded feature keyline set (step 4), then train and measure the classifier's performance based on the updated feature vectors (step 5). The number of iterations in the feature keyline set expansion algorithm depends on the number of feature candidate keylines in the train sets (considered in multiple runs of the 3-fold cross validation process as described in Section~\ref{sec:method_expansion}).
In other words, for a particular instantiation of the cross-validation step, the number of iterations depends on the number of positive samples in the train set (which is randomly sampled for each instance of cross-validation).
Due to such variability, for Algorithm 1 we observed 30-33 iterations for congestion, 23-25 for auto, 38-40 for transit, and 16-18 for bike feature respectively.

\begin{figure}[!htb]
\subfigure[Congestion : $\mathbf{K}^{(c)}$]{\label{fig:congestion_cv}
\includegraphics[scale=.53]{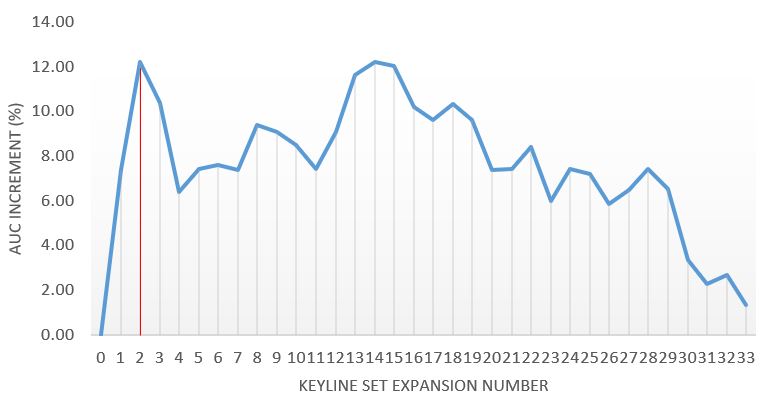}\hspace{1em}
}
\subfigure[Auto : $\mathbf{K}^{(a)}$]{\label{fig:auto_cv}
\includegraphics[scale=.55]{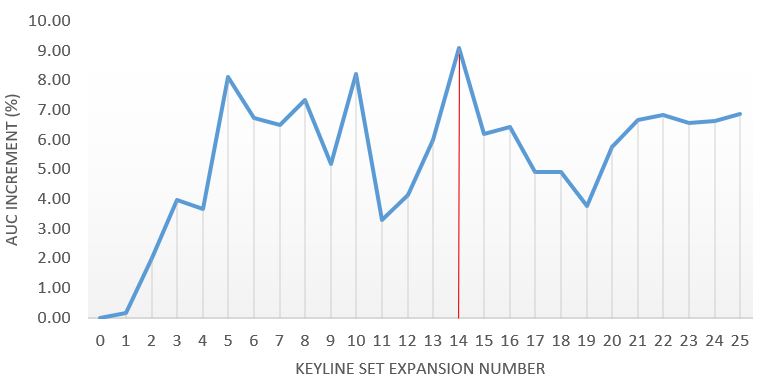}\hspace{1em}
}
\subfigure[Transit : $\mathbf{K}^{(t)}$]{\label{fig:transit_cv}
\includegraphics[scale=.55]{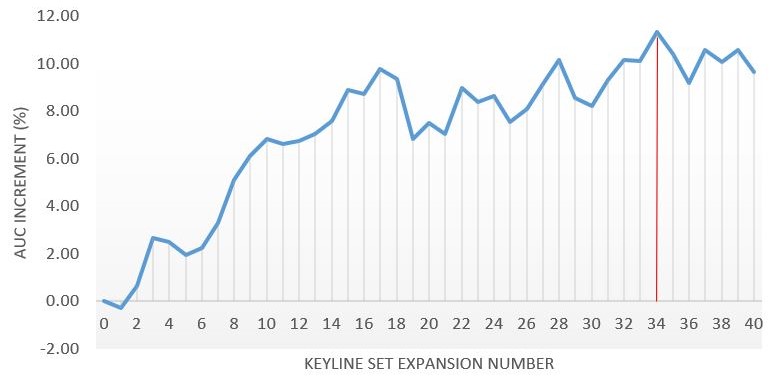}\hspace{1em}
}
\subfigure[Bike : $\mathbf{K}^{(b)}$]{\label{fig:bike_cv}
\includegraphics[scale=.57]{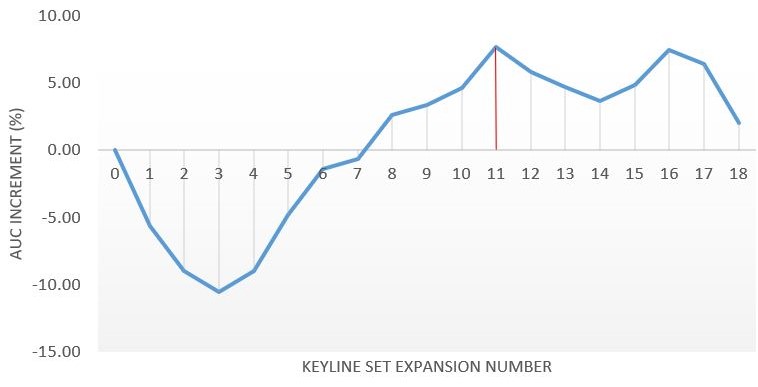}\hspace{1em}
}
\caption{Optimal keyline set expansion for: (a) congestion, (b) auto, (c) transit, and (d) bike. The keyline set expansion number (X-axis) refers to the iteration number, and percentage increment in AUC (Y-axis) corresponds to the mean model performance measured at each iteration (expansion) using cross-validation} \label{fig:keyline_expansion_cv}
\end{figure}

At each $e$ ($\geq$ 1), the percentage increment in the model performance metric (\emph{i.e.}, average validation AUC) is computed with respect to $e=0$ (\emph{i.e.}, where the model is trained with only anchor text-based features $f_{initial}^{(c)}$, $f_{initial}^{(a)}$, $f_{initial}^{(t)}$, and  $f_{initial}^{(b)}$). Figure~\ref{fig:keyline_expansion_cv} plots the performance metric increment graph with incremental expansion of the feature keyline set. \textcolor{black}{The plots highlight the optimal number of representative keylines per typology (in addition to their respective anchor texts) that results in the highest increment in the model performance (\emph{i.e.,} based on validation sets). As highlighted in the figure, it is observed that the congestion typology classifier performs best at $e = 2$ (Figure~\ref{fig:keyline_expansion_cv}(a)). This means, three representative keylines indicative of congestion in cities (added from
$sorted\_candidates^{(c)}$) including $anchor^{(c)}$ in $\mathbf{K}_{opt}^{(c)}$ (as explained in Algorithm 1) is found optimal for representing the congestion feature $f^{(c)}$. As the congestion feature of a city indicates the degree of relevant information pertaining to congestion present in its Wikipedia page; the congestion plot suggests that three representative keylines are optimal in retrieving useful signals from city Wikipedia pages (pertaining to congestion typology) for an effective typology classification.} Similarly, for auto, transit, and bike, best performances are noticed at $e$ = $14$, $34$, and $11$ respectively. We denote the optimal feature keyline sets for congestion, auto, transit, and bike as $\mathbf{K}_{opt}^{(c)}$, $\mathbf{K}_{opt}^{(a)}$, $\mathbf{K}_{opt}^{(t)}$, and  $\mathbf{K}_{opt}^{(b)}$ such that $|\mathbf{K}_{opt}^{(c)}|$ = 3, $|\mathbf{K}_{opt}^{(a)}|$ = 15, $|\mathbf{K}_{opt}^{(t)}|$ = 35, and 
$|\mathbf{K}_{opt}^{(b)}|$ = 12 respectively. Similarly, features computed based on respective optimal feature keyline sets are denoted as $f_{opt}^{(c)}$, $f_{opt}^{(a)}$, $f_{opt}^{(t)}$, and  $f_{opt}^{(b)}$; as examples Table~\ref{tab:final_model_features} provides values of these features for selected cities (as listed in Table~\ref{tab:initial_model_features}).

\begin{table}[h]
    \centering
    \caption{Keyline feature values (obtained using the optimal feature keyline sets) for example cities in the typology data}
    \begin{tabular}{|l|c|c|c|c|c|}
    \hline
    City name & $f_{opt}^{(c)}$ & $f_{opt}^{(a)}$ & $f_{opt}^{(t)}$ &  $f_{opt}^{(b)}$ & City typology\\
    & & & & &\\
     \hline
       Dhaka, Bangladesh	&	0.781	&	0.666	&	0.649	&	0.658	&	Congestion\\
       Dubai, United Arab Emirates	&	0.552	&	0.716	&	0.621	&	0.615	&	Auto-heavy\\
       Amsterdam, Netherlands	&	0.554	&	0.601	&	0.911	&	0.776	&	Transit-heavy\\
       Changchun, China	&	0.510	&	0.671	&	0.609	&	0.717	&	Bike-friendly\\
    \hline
    \end{tabular}
    \label{tab:final_model_features}
\end{table}

Table~\ref{tab:optimal_keylines_example} provides some examples of expanded keylines from each optimal feature keyline sets. The relevance of most of the expanded feature keylines in indicating the underlying meaning of the typology is worthy of note. For example, the transit keyline \textit{'every major street in the city is served by at least one bus route'} indicates the extensive use of bus transit in the city. In addition, the textual diversity in the expanded feature keylines is a positive indication that our model is able to find relevant keylines given the anchor text despite the paraphrasing complexity in hundreds of Wikipedia pages (spanning multiple countries and human editors). Moreover, while processing lines from a city’s Wikipedia page for predicting the typology, we compute the maximum similarity across a set of keylines (as explained in Section~\ref{sec:keyline_feature_computation}); this limits the chances of incorrect inference even when a small fraction of keylines is noisy or not relevant. 

%% EXAMPLES OPTIMAL EXPANDED KEYLINES %%
\begin{table}[h]
    \centering
    %\scalebox{0.9}{
     \caption{Examples of keylines (extracted from city Wikipedia pages) in the optimal feature keyline sets} 
    \begin{tabular}{c|p{3cm} p{3.2cm} p{2.8cm} p{3cm}}
    \hline
	&	Auto keylines	&	Congestion keylines	&	Bike keylines	&	Transit keylines\\
	\hline\\
Anchor text	&	Most people in the city use cars	&	The city has heavy traffic congestion	&	Many people in the city use bike or cycle	&	Most people in the city use public transit like bus and metro\\
\hline\\
\multirow{2}{*}{Expanded keylines} &	The area has a number of freeways to transport \;\;\;\;\; people by car	&	Uncontrolled urban sprawl has                                            challenged the city infrastructure, producing  heavy traffic congestion	&	 It is possible to                                            cycle to most parts of the city	& Almost half of all journeys in the metropolitan area are made on                                         public transport\\
                                        \cline{2-5}\\

       	                            &	Car sharing is available to residents of the city and some inner suburbs	&	Chronic traffic congestion, and a sudden and prolonged surge in crime have become perennial problems	&	Cycling has seen a resurgence in popularity due to the emergence of a large number of dockless app-based bicycle sharing systems	&	Every major street in the city is served by at least one bus route\\
    \hline
    \end{tabular}
    \label{tab:optimal_keylines_example}
\end{table}

Based on the graphs plotted in Figure~\ref{fig:keyline_expansion_cv}, the model performance improves when a representative set of keylines (indicating the typology-based feature) is considered compared to anchor text alone, even though the additional keylines are based on the same underlying meaning as the anchor text. This is because various keylines that are extracted from city Wikipedia pages are not just semantically similar to the intuitive meaning of the typology, they represent city-based factors and elements associated with the typology as well (as can be seen in Table~\ref{tab:optimal_keylines_example}). Examples include inadequate city infrastructure leading to congestion, availability of sufficient freeways facilitating auto use, or emergence of app-based bicycle sharing services in popularizing cycling in the city. Having a representative set of feature keylines (the full list of expanded keylines is available in Zenodo \citealp{zenodo}) %\href{https://docs.google.com/spreadsheets/d/1DlbbUmJbL1J2vudwZ-KXDvgRkPkek71Oa8SzuYxL9Gc/edit?usp=sharing}{https://docs.google.com/spreadsheets/d/1DlbbUmJbL1J2vudwZ-KXDvgRkPkek71Oa8SzuYxL9Gc/edit?usp=sharing}} 
casts a wider net for retrieving useful signals from the city Wikipedia pages that indicate or represent the typology of interest. Moreover, addition of keylines beyond optimal expansion (as highlighted in the plots in Figure~\ref{fig:keyline_expansion_cv}) does not necessarily improve the performance further; this is possibly due to introduction of irrelevant lines or noise in the keyline set.

%% TYPOLOGY CLASSIFIER RESULTS %%
\subsection{Typology classification results} \label{sec:performance}
We discuss the performance of the four typology classifiers developed in our study \emph{i.e.}, congestion, auto-heavy, transit-heavy and bike-friendly. 
\textcolor{black}{Since the proposed methodology focuses on extracting useful data from Wikipedia to define city features, we examine model performances using feature combinations based on different choices of feature keyline sets (anchor texts, optimal keyline sets etc.) and features from different Wikipedia components (i.e., numeric and textual features). The focus of the analyses is to provide an insightful performance comparison across different feature sets and data extracted from Wikipedia in multiple typology classification tasks while maintaining model interpretability.} The input features used in the typology classification models constitute the following: features from Wikipedia textual component only (\emph{i.e.,} $f^{(c)}$, $f^{(a)}$, $f^{(t)}$, and $f^{(b)}$); features from Wikipedia numeric component only (\emph{i.e.}, $f^{(density)}$), and both textual and numeric components. 
As described in Section~\ref{sec:keyline_feature_computation}, the model features $f^{(c)}$, $f^{(a)}$, $f^{(t)}$, and $f^{(b)}$ are computed based on respective feature keyline sets.  
\textcolor{black}{Therefore, in addition to aforementioned sets of input features, we use different combinations of feature keyline sets to define the model features along with considering different subset of features related to typology classes in each typology classification task. In general, the problem of finding the optimal subset of features (from an available set of features) is combinatorial in nature and has exponential complexity; but there exist greedy approaches and heuristics to tackle this problem for a large number of features (\citealp{chandrashekar2014survey}). Such optimal feature subset selection is not the primary focus in our study, but since the number of features in our setup $(= 5)$ is low, we were able to conduct experiments around different feature combinations to derive useful insights.}

The keyline sets for congestion, auto, transit, and bike include: singleton sets with anchor texts only ($\mathbf{K}^{(c)}_{initial}$, $\mathbf{K}^{(a)}_{initial}$, $\mathbf{K}^{(t)}_{initial}$, $\mathbf{K}^{(b)}_{initial}$); optimal sets where keylines are optimally expanded using keyline set expansion method ($\mathbf{K}^{(c)}_{opt}$, $\mathbf{K}^{(a)}_{opt}$, $\mathbf{K}^{(t)}_{opt}$, $\mathbf{K}^{(b)}_{opt}$ obtained in Section~\ref{sec:keyline set expansion results}); and full sets where keylines are expanded to the fullest \emph{i.e.}, using all candidate keylines (we denote them as $\mathbf{K}^{(c)}_{all}$, $\mathbf{K}^{(a)}_{all}$, $\mathbf{K}^{(t)}_{all}$, $\mathbf{K}^{(b)}_{all}$). The notations used for features are based on the notations of feature keyline sets. For example, the congestion feature $f^{(c)}$ computed using $\mathbf{K}^{(c)}_{initial}$, $\mathbf{K}^{(c)}_{opt}$, and $\mathbf{K}^{(c)}_{all}$ is denoted as $f_{initial}^{(c)}$, $f_{opt}^{(c)}$, and $f_{all}^{(c)}$ respectively. Similar notations are used to denote auto, transit, and bike features. 

\textcolor{black}{Therefore, using multiple combinations of the above mentioned features, we compute 21 feature subsets for each typology classification task 
(as shown in \Cref{tab:C_model_output,tab:A_model_output,tab:T_model_output,tab:B_model_output})}. For each of these choices, the classifier is trained using the train set (197 cities) and we study the test set metrics (using 85 cities in the test set) for the choice of feature combinations. \Cref{tab:C_model_output,tab:A_model_output,tab:T_model_output,tab:B_model_output} report
the test set performance metrics of the typology classifiers along with details of the feature subsets used as inputs in the models.
In each case, percentage lift in the test set AUC score (obtained using corresponding model features) are reported; lift is calculated with respect to the model trained using only anchor text-based features.

%%%% CONGESTION MODEL %%%%%%
\begin{table}[h]
    \centering
    \caption{Congestion typology prediction model output for multiple input feature subsets}
    {\color{black}\begin{tabular}{l c c | l c c }
    \hline
      Model features  & AUC (test)  & Lift 
      & Model features  & AUC (test)  & Lift\\
     \\[-1em]
     \hline
        ${f}^{(c)}_{initial}$, ${f}^{(a)}_{initial}$, ${f}^{(t)}_{initial}$, ${f}^{(b)}_{initial}$  &  0.53   &  0.0\% &
        $f^{(density)}$ & 0.81 & +52.8\% \\
        &     &   & &  &\\
        \hline
       ${f}^{(c)}_{opt}$, ${f}^{(a)}_{initial}$, ${f}^{(t)}_{initial}$, ${f}^{(b)}_{initial}$  &  0.61  &  +15.1\% & 
        ${f}^{(c)}_{opt}$, $f^{(density)}$ &   0.82   &  +54.7\%   \\
         &     &   & &  &\\
       \hline
        ${f}_{all}^{(c)}$, ${f}^{(a)}_{initial}$, ${f}^{(t)}_{initial}$, ${f}^{(b)}_{initial}$ & 0.56 & +5.7\% & 
        ${f}^{(c)}_{opt}$, ${f}^{(a)}_{initial}$, ${f}^{(t)}_{initial}$, ${f}^{(b)}_{initial}$,$f^{(density)}$  & 0.83 & +56.6\%   \\
         &     &   & &  &\\
       \hline
        ${f}^{(c)}_{opt}$ & 0.61 & +15.1\% & ${f}^{(c)}_{opt}$, ${f}^{(a)}_{opt}$, $f^{(density)}$ & 0.85 & +60.4\%\\
         &     &   & &  &\\
        \hline
       ${f}^{(c)}_{opt}$, ${f}^{(a)}_{opt}$ & 0.75 & +41.5\% & ${f}^{(c)}_{opt}$, ${f}^{(t)}_{opt}$, $f^{(density)}$ & 0.85 & +60.4\% \\
       &     &   & &  &\\
       \hline
       ${f}^{(c)}_{opt}$, ${f}^{(t)}_{opt}$ & 0.68  & +28.3\% & ${f}^{(c)}_{opt}$, ${f}^{(b)}_{opt}$, $f^{(density)}$ & 0.83 & +56.6\% \\
        &     &   & &  &\\
       \hline
       ${f}^{(c)}_{opt}$, ${f}^{(b)}_{opt}$ & 0.62  & +17.0\% & \textbf{${f}^{(c)}_{opt}$, ${f}^{(a)}_{opt}$, ${f}^{(t)}_{opt}$, $f^{(density)}$} & \textbf{0.87} & \textbf{+64.2\%} \\
        &     &   & &  &\\
       \hline
       ${f}^{(c)}_{opt}$, ${f}^{(a)}_{opt}$, ${f}^{(t)}_{opt}$ & 0.74  & +39.6\% & ${f}^{(c)}_{opt}$, ${f}^{(a)}_{opt}$, ${f}^{(b)}_{opt}$, $f^{(density)}$ & 0.86 & +62.3\% \\
        &     &   & &  &\\
       \hline
       ${f}^{(c)}_{opt}$, ${f}^{(a)}_{opt}$, ${f}^{(b)}_{opt}$ & 0.75  & +41.5\% & ${f}^{(c)}_{opt}$, ${f}^{(t)}_{opt}$, ${f}^{(b)}_{opt}$, $f^{(density)}$ & 0.85 & +60.4\% \\
        &     &   & &  &\\
       \hline
        ${f}^{(c)}_{opt}$, ${f}^{(t)}_{opt}$, ${f}^{(b)}_{opt}$ & 0.67  & +26.4\% & \textbf{${f}^{(c)}_{opt}$, ${f}^{(a)}_{opt}$, ${f}^{(t)}_{opt}$, ${f}^{(b)}_{opt}$, $f^{(density)}$} & \textbf{0.87} & \textbf{+64.2\%} \\
         &     &   & &  &\\
        \hline
       ${f}^{(c)}_{opt}$, ${f}^{(a)}_{opt}$, ${f}^{(t)}_{opt}$, ${f}^{(b)}_{opt}$ & 0.74 & +39.6\% &      &    & \\
        &     &   & &  &\\
    \hline
    \end{tabular}}
    \label{tab:C_model_output}
\end{table}

%%% AUTO MODEL %%%%
\begin{table}[h]
    \centering
    \caption{Auto-heavy typology prediction model output for multiple input feature subsets}
    {\color{black}\begin{tabular}{l c c | l c c }
    \hline
      Model features  & AUC (test)  & Lift 
      & Model features  & AUC (test)  & Lift\\
      \\[-1em]
     \hline
        ${f}^{(a)}_{initial}$, ${f}^{(c)}_{initial}$, ${f}^{(t)}_{initial}$, ${f}^{(b)}_{initial}$  &  0.55   &  0.0\% &
        $f^{(density)}$ & 0.79 & +43.6\% \\
         &     &   & &  &\\
        \hline
       ${f}^{(a)}_{opt}$, ${f}^{(c)}_{initial}$, ${f}^{(t)}_{initial}$, ${f}^{(b)}_{initial}$  &  0.75  &  +36.4\% & 
        ${f}^{(a)}_{opt}$, $f^{(density)}$ &   0.83   &  +50.9\%   \\
         &     &   & &  &\\
       \hline
        ${f}_{all}^{(a)}$, ${f}^{(c)}_{initial}$, ${f}^{(t)}_{initial}$, ${f}^{(b)}_{initial}$ & 0.74 & +34.5\% & 
        ${f}^{(a)}_{opt}$, ${f}^{(c)}_{initial}$, ${f}^{(t)}_{initial}$, ${f}^{(b)}_{initial}$,$f^{(density)}$  & 0.82 & +49.1\%   \\
         &     &   & &  &\\
       \hline
        ${f}^{(a)}_{opt}$ & 0.68 & +23.6\% & ${f}^{(a)}_{opt}$, ${f}^{(c)}_{opt}$, $f^{(density)}$ & 0.83 & +50.9\% \\
        &     &   & &  &\\
        \hline
       ${f}^{(a)}_{opt}$, ${f}^{(c)}_{opt}$ & 0.72 & +30.9\% & ${f}^{(a)}_{opt}$, ${f}^{(t)}_{opt}$, $f^{(density)}$ & 0.85 & +54.5\% \\
        &     &   & &  &\\
       \hline
       ${f}^{(a)}_{opt}$, ${f}^{(t)}_{opt}$ & 0.73  & +32.7\% & ${f}^{(a)}_{opt}$, ${f}^{(b)}_{opt}$, $f^{(density)}$ & 0.84 & +52.7\% \\
        &     &   & &  &\\
       \hline
       ${f}^{(a)}_{opt}$, ${f}^{(b)}_{opt}$ & 0.75  & +36.4\% & ${f}^{(a)}_{opt}$, ${f}^{(c)}_{opt}$, ${f}^{(t)}_{opt}$, $f^{(density)}$ & 0.85 & +54.5\% \\
        &     &   & &  &\\
       \hline
       ${f}^{(a)}_{opt}$, ${f}^{(c)}_{opt}$, ${f}^{(t)}_{opt}$ & 0.73  & +32.7\% & ${f}^{(a)}_{opt}$, ${f}^{(c)}_{opt}$, ${f}^{(b)}_{opt}$, $f^{(density)}$ & 0.85 & +54.5\% \\
        &     &   & &  &\\
       \hline
       ${f}^{(a)}_{opt}$, ${f}^{(c)}_{opt}$, ${f}^{(b)}_{opt}$ & 0.76  & +38.2\% & ${f}^{(a)}_{opt}$, ${f}^{(t)}_{opt}$, ${f}^{(b)}_{opt}$, $f^{(density)}$ & 0.85 & +54.5\% \\
       &     &   & &  &\\
       \hline
        ${f}^{(a)}_{opt}$, ${f}^{(t)}_{opt}$, ${f}^{(b)}_{opt}$ & 0.75  & +36.4\% & \textbf{${f}^{(a)}_{opt}$, ${f}^{(c)}_{opt}$, ${f}^{(t)}_{opt}$, ${f}^{(b)}_{opt}$, $f^{(density)}$} & \textbf{0.86} & \textbf{+56.4\%} \\
         &     &   & &  &\\
        \hline
       ${f}^{(a)}_{opt}$, ${f}^{(c)}_{opt}$, ${f}^{(t)}_{opt}$, ${f}^{(b)}_{opt}$ & 0.75 & +36.4\% &      &    & \\
        &     &   & &  &\\
    \hline
    \end{tabular}}
    \label{tab:A_model_output}
\end{table}

%%% TRANSIT MODEL %%%
\begin{table}[h]
    \centering
    \caption{Transit-heavy typology prediction model output for multiple input feature subsets}
    {\color{black}\begin{tabular}{l c c | l c c }
    \hline
      Model features  & AUC (test)  & Lift 
      & Model features  & AUC (test)  & Lift\\
      \\[-1em]
     \hline
        ${f}^{(t)}_{initial}$, ${f}^{(a)}_{initial}$, ${f}^{(c)}_{initial}$, ${f}^{(b)}_{initial}$  &  0.56   &  0.0\% &
        $f^{(density)}$ & 0.5 & -10.7\% \\
         &     &   & &  &\\
        \hline
       ${f}^{(t)}_{opt}$, ${f}^{(a)}_{initial}$, ${f}^{(c)}_{initial}$, ${f}^{(b)}_{initial}$  &  0.59  &  +5.4\% & 
        ${f}^{(t)}_{opt}$, $f^{(density)}$ &   0.58   &  +3.6\%   \\
         &     &   & &  &\\
       \hline
        ${f}_{all}^{(t)}$, ${f}^{(a)}_{initial}$, ${f}^{(c)}_{initial}$, ${f}^{(b)}_{initial}$ & 0.55 & -1.8\% & 
        ${f}^{(t)}_{opt}$, ${f}^{(a)}_{initial}$, ${f}^{(c)}_{initial}$, ${f}^{(b)}_{initial}$,$f^{(density)}$  & 0.58 & +3.6\%   \\
         &     &   & &  &\\
       \hline
        ${f}^{(t)}_{opt}$ & 0.59 & +5.4\% & ${f}^{(t)}_{opt}$, ${f}^{(a)}_{opt}$, $f^{(density)}$ & 0.58 & +3.6\%  \\
        &     &   & &  &\\
        \hline
       ${f}^{(t)}_{opt}$, ${f}^{(a)}_{opt}$ & 0.6 & +7.1\% & ${f}^{(t)}_{opt}$, ${f}^{(c)}_{opt}$, $f^{(density)}$ & 0.6 & +7.1\% \\
        &     &   & &  &\\
       \hline
       \textbf{${f}^{(t)}_{opt}$, ${f}^{(c)}_{opt}$} & \textbf{0.61}  & \textbf{+8.9\%} & ${f}^{(t)}_{opt}$, ${f}^{(b)}_{opt}$, $f^{(density)}$ & 0.58 & +3.6\% \\
        &     &   & &  &\\
       \hline
       ${f}^{(t)}_{opt}$, ${f}^{(b)}_{opt}$ & 0.59  & +5.4\% & ${f}^{(t)}_{opt}$, ${f}^{(a)}_{opt}$, ${f}^{(c)}_{opt}$, $f^{(density)}$ & 0.59 & +5.4\% \\
        &     &   & &  &\\
       \hline
       ${f}^{(t)}_{opt}$, ${f}^{(a)}_{opt}$, ${f}^{(c)}_{opt}$ & 0.6  & +7.1\% & ${f}^{(t)}_{opt}$, ${f}^{(a)}_{opt}$, ${f}^{(b)}_{opt}$, $f^{(density)}$ & 0.58 & +3.6\% \\
        &     &   & &  &\\
       \hline
       ${f}^{(t)}_{opt}$, ${f}^{(a)}_{opt}$, ${f}^{(b)}_{opt}$ & 0.58 & +3.6\% & ${f}^{(t)}_{opt}$, ${f}^{(c)}_{opt}$, ${f}^{(b)}_{opt}$, $f^{(density)}$ & 0.59 & +5.4\% \\
       &     &   & &  &\\
       \hline
        ${f}^{(t)}_{opt}$, ${f}^{(c)}_{opt}$, ${f}^{(b)}_{opt}$ & 0.6  & +7.1\% & \textbf{${f}^{(t)}_{opt}$, ${f}^{(a)}_{opt}$, ${f}^{(c)}_{opt}$, ${f}^{(b)}_{opt}$, $f^{(density)}$} & 0.58 & +3.6\% \\
         &     &   & &  &\\
        \hline
       ${f}^{(t)}_{opt}$, ${f}^{(a)}_{opt}$, ${f}^{(c)}_{opt}$, ${f}^{(b)}_{opt}$ & 0.59 & +5.4\% &      &    & \\
        &     &   & &  &\\
    \hline
    \end{tabular}}
    \label{tab:T_model_output}
\end{table}

%%% BIKE MODEL %%%
\begin{table}[h]
    \centering
    \caption{Bike-friendly typology prediction model output for multiple input feature subsets}
    {\color{black}\begin{tabular}{l c c | l c c }
    \hline
      Model features  & AUC (test)  & Lift 
      & Model features  & AUC (test)  & Lift\\
      \\[-1em]
     \hline
        ${f}^{(b)}_{initial}$, ${f}^{(c)}_{initial}$, ${f}^{(t)}_{initial}$, ${f}^{(a)}_{initial}$  &  0.5  &  0.0\% &
        $f^{(density)}$ & 0.85 & +70.0\% \\
         &     &   & &  &\\
        \hline
       ${f}^{(b)}_{opt}$, ${f}^{(c)}_{initial}$, ${f}^{(t)}_{initial}$, ${f}^{(a)}_{initial}$  &  0.81  &  +62.0\% & 
        \textbf{${f}^{(b)}_{opt}$, $f^{(density)}$} &  \textbf{0.94}  &  \textbf{+88.0\%}   \\
         &     &   & &  &\\
       \hline
        ${f}_{all}^{(b)}$, ${f}^{(c)}_{initial}$, ${f}^{(t)}_{initial}$, ${f}^{(a)}_{initial}$ &  0.8  &  +60.0\% & 
        ${f}^{(b)}_{opt}$, ${f}^{(c)}_{initial}$, ${f}^{(t)}_{initial}$, ${f}^{(a)}_{initial}$,$f^{(density)}$  &  0.81  &  +62.0\%  \\
         &     &   & &  &\\
       \hline
        ${f}^{(b)}_{opt}$ &  0.86  &  +72.0\% & ${f}^{(b)}_{opt}$, ${f}^{(c)}_{opt}$, $f^{(density)}$ &  0.86  &  +72.0\%  \\
        &     &   & &  &\\
        \hline
       ${f}^{(b)}_{opt}$, ${f}^{(c)}_{opt}$ &  0.77  &  +54.0\% & ${f}^{(b)}_{opt}$, ${f}^{(t)}_{opt}$, $f^{(density)}$ &  0.92  &  +84.0\% \\
        &     &   & &  &\\
       \hline
      ${f}^{(b)}_{opt}$, ${f}^{(t)}_{opt}$ & 0.85  & +70.0\% & \textbf{${f}^{(b)}_{opt}$, ${f}^{(a)}_{opt}$, $f^{(density)}$} &  \textbf{0.94}  & \textbf{+88.0\%} \\
        &     &   & &  &\\
       \hline
       ${f}^{(b)}_{opt}$, ${f}^{(a)}_{opt}$ & 0.86  & +72.0\% & ${f}^{(b)}_{opt}$, ${f}^{(c)}_{opt}$, ${f}^{(t)}_{opt}$, $f^{(density)}$ & 0.89  & +78.0\% \\
        &     &   & &  &\\
       \hline
       ${f}^{(b)}_{opt}$, ${f}^{(c)}_{opt}$, ${f}^{(t)}_{opt}$ & 0.75  & +50.0\% & ${f}^{(b)}_{opt}$, ${f}^{(c)}_{opt}$, ${f}^{(a)}_{opt}$, $f^{(density)}$ & 0.87 & +74.0\% \\
        &     &   & &  &\\
       \hline
       ${f}^{(b)}_{opt}$, ${f}^{(c)}_{opt}$, ${f}^{(a)}_{opt}$ & 0.78 & +56.0\% & ${f}^{(b)}_{opt}$, ${f}^{(t)}_{opt}$, ${f}^{(a)}_{opt}$, $f^{(density)}$ & 0.93 & +86.0\% \\
       &     &   & &  &\\
       \hline
        ${f}^{(b)}_{opt}$, ${f}^{(t)}_{opt}$, ${f}^{(a)}_{opt}$ & 0.85  & +70.0\% & ${f}^{(b)}_{opt}$, ${f}^{(c)}_{opt}$, ${f}^{(t)}_{opt}$, ${f}^{(a)}_{opt}$, $f^{(density)}$ & 0.92 & +84.0\% \\
         &     &   & &  &\\
        \hline
       ${f}^{(b)}_{opt}$, ${f}^{(c)}_{opt}$, ${f}^{(t)}_{opt}$, ${f}^{(a)}_{opt}$ & 0.76 & +52.0\% &      &    & \\
        &     &   & &  &\\
    \hline
    \end{tabular}}
    \label{tab:B_model_output}
\end{table}

\newpage
Based on the results of the typology classifiers on the test set, it is observed that, \textcolor{black}{in most of the cases, including features from other classes (different subsets in each case) provide better performance compared to using only the respective class (single) feature. Significant improvements are observed in congestion classifier (with $f_{opt}^{(c)},f_{opt}^{(a)}$,and $f_{opt}^{(b)}$) and auto-heavy classifier (with $f_{opt}^{(a)},f_{opt}^{(c)},$ and $f_{opt}^{(b)}$) with a slight uplift in transit-heavy classifier (with $f_{opt}^{(t)}$ and $f_{opt}^{(c)}$), whereas bike-friendly classifier show similar performance with single feature $f_{opt}^{(b)}$ and with the feature subset $f_{opt}^{(b)}$ and $f_{opt}^{(a)}$. 
Moreover, compared to using only anchor text-based features, the models for congestion, auto-heavy, and bike-friendly typology prediction show the highest lift (64-88\% lift in AUC) when features from both numeric and textual components are used, where the textual features are computed based on their optimal feature keylines (highlighted in bold in \Cref{tab:C_model_output,tab:A_model_output,tab:T_model_output,tab:B_model_output}). In the congestion classifier, this is observed when textual features from all the four classes are considered (similar performance noticed with three classes as well i.e., $f_{opt}^{(c)},f_{opt}^{(a)},f_{opt}^{(t)}$). In the auto-heavy classifier, textual features from all the four classes give the highest lift, whereas in the bike-friendly classifier, this corresponds to the case when the feature subset from two typologies (bike and auto related) are used \emph{i.e.,} $f_{opt}^{(b)}$ and $f_{opt}^{(a)}$ (using only respective class feature $f_{opt}^{(b)}$ also provide similar observations). On the other hand, in the transit-heavy typology prediction model, the highest lift in performance is achieved with a subset of transit and congestion related (textual) features only (\emph{i.e.,} $f_{opt}^{(t)}$ and $f_{opt}^{(c)}$). 
However, even with a 9\% improvement in the performance, the transit-heavy prediction model can be improved further with inclusion of additional city-based features,
as it seems that current Wikipedia pages may not be providing sufficient information for transit-heavy typology prediction for cities.} 

\textcolor{black}{The performances of the four separate typology classifiers using our proposed approach indicate that Wikipedia alone may not be equally sufficient for all types of city typology prediction. However, as the goal of the study is to investigate the use of Wikipedia as an effective data source in transportation, our experiments find Wikipedia articles to be informative about transportation-based typology indicators for cities. Moreover, while our study evaluates the appropriateness of using Wikipedia alone for different typology classification tasks, further enhancing the model performances with additional innovative features (e.g., features extracted from city images or satellite network structures that are also accessible for a large number of cities) can be a promising direction for future research.}
%This applies to other typology classification models as well, where further enhancing the model performance with extra features (from Wikipedia and/or other data sources) can be a promising future research direction. 

For the models with the highest test set AUC scores, we report the input feature coefficients (including intercepts) and additional performance metrics (\emph{i.e.}, classification scores as explained in Section~\ref{sec:metrics}) in 
Table~\ref{tab:model_coeff}. 
\textcolor{black}{Note that, for cases where two different feature subsets provide similar results (\emph{i.e.,} highest performance), we choose the subset with more classes to get the model output (in Table~\ref{tab:model_coeff}) for a better insight on the effects of different typology features in different city typology classification tasks.}
%get the input feature coefficients (including intercepts) and calculate additional performance metrics (\emph{i.e.}, classification scores as explained in Section~\ref{sec:metrics}); values are reported in Table~\ref{tab:model_coeff}.
Features with positive coefficient influence the probability of the event (typology prediction in our case) in a positive way and vice-versa. It is worth highlighting  that in each typology classification model, features corresponding to the typology prediction have positive coefficients. For example, the auto-heavy classifier has a positive coefficient for $f^{(a)}$; this implies if there is indicative information in Wikipedia regarding high usage of automobiles in a city, there is a high chance that the city is predicted as an auto-heavy type city. 
\textcolor{black}{As we build upon the notion of distinct typologies based on \cite{okeMIT}, consistency in the (observed) effects of the keyline features across different typology classifiers further validates the relevance of the typology-based keylines obtained using our proposed method.} Moreover, the population density feature in the congestion classification model has a positive coefficient, this justifies the relation between population growth and traffic congestion in cities. 

%The optimal decision thresholds for each model are used to obtain the classification metrics as shown in Table~\ref{tab:model_coeff}. for each model

It is interesting to note the reasonably high classification scores for congestion, auto-heavy, and bike-friendly typology prediction models reflecting their generalization
capabilities; this is important
when the models are be used on new and unseen data. However, we must also note
that the performance of the model for transit-heavy typology prediction can be further enhanced with supplementary features supporting the typology, and additional data for training the model.
Nonetheless, the classification results are rather encouraging indicating the effectiveness of Wikipedia as a data source for predicting city typologies.

%%% MODELS WITH HIGHEST TEST SET AUC FEATURES %%%
\begin{table}[h]
\centering
    \caption{Feature coefficients and classification scores of the best performing city typology prediction models}
    \scalebox{0.98}{
    \color{black}{\begin{tabular}{l|c c c c c c c c c c c}
    \hline
Model (task)	&	Intercept	&	$f^{(c)}$	&	 $f^{(a)}$	&	$f^{(t)}$	& $f^{(b)}$	&	$f^{(density)}$	&	Accuracy	&	Precision	&	Recall	&	F1-Score\\
\\[-1em]
     \hline
     \\[-1em]
Congestion prediction	&	-0.699	&	0.119	&	-0.184	&	-0.274	&	-0.064	&	0.537	&	0.79	&	0.85	&	0.79	&	0.80\\
\\[-1em]
Auto-heavy prediction	&	-1.215	&	-0.001	&	0.047	&	-0.024	&	-0.013	&	-0.032	&	0.82	&	0.84	&	0.82	&	0.83\\
\\[-1em]
Transit-heavy prediction	&	-3.021	&	-0.44	&	0.000	&	3.667	&	0.000	&	0.000	&	0.6	&	0.6	&	0.6	&	0.6\\
\\[-1em]
Bike-friendly prediction	&	-4.822	&	0.000	&	-4.593	&	0.000	&	10.052	&	-27.631	&	0.88	&	0.94	&	0.88	&	0.90\\
    \hline
    \end{tabular}}}
    \label{tab:model_coeff}
\end{table}

%% INSIGHTS %%
\subsection{Insights and applications}\label{sec:insights}
The method proposed in our study for extracting useful information from a city's Wikipedia page for the purpose of understanding the transportation-based typology of the city assumes that Wikipedia pages have relevant information pertaining to such typologies. To help us better understand how well the typology classification models in our study perform in this context, we provide a few examples in Table~\ref{tab:best_similar_text_examples}. This table lists a few cities (from the test set) along with the Wikipedia text of each city; this refers to the text (line in the city's Wikipedia page) that form the feature pertaining to the predicted typology of the city (obtained using the best performing typology model as in Table~\ref{tab:model_coeff}). For example, the Wikipedia text for the city San Diego in Table~\ref{tab:best_similar_text_examples} corresponds to its auto feature ($f^{(a)} \Leftrightarrow f_{opt}^{(a)}$) used in the auto-heavy classification model, and this model predicts San Diego as an auto-heavy city (based on the optimal threshold described in \ref{sec:performance}). Other cities in the table can be interpreted in a similar manner.
In view of the results, it is evident that Wikipedia can be a useful source of mobility or transport related information for cities
and a relevant data source for typology prediction tasks.\\

%% TYPOLOGY INFORMATION IN WIKIPEDIA PAGES %%
\begin{table}[h]
    \centering
    \caption{Examples of (textual) information obtained from city Wikipedia pages pertaining to the predicted city typology}
    \begin{tabular}{c|p{3cm} p{3.2cm} p{2.7cm} p{3cm}}
    \hline
	Predicted typology &	Auto	&	Congestion	&	Bike 	&	Transit\\
	\hline\\
\multirow{2}{*}{City:Wikipedia text} &	San Diego (U.S.A): \textit{With the automobile being the primary means of transportation for over 80 percent of residents , San Diego is served by a network of freeways and highways}	&  Kabul (Afghanistan): \textit{The steep population rise in the 21st century has caused major congestion problems for the city roads} & Shijiazhuang (China): \textit{Thousands of cyclists use the city each day and often there are more cyclists waiting at a crossroad than cars} & Durban (South Africa):  \textit{Several companies run long - distance bus services from Durban to the other cities in South Africa} \\
 \cline{2-5}\\
& Los Angeles (U.S.A): \textit{The city and the rest of the Los Angeles metropolitan area are served by an extensive network of freeways and highways} & Delhi (India): \textit{Delhi's rapid rate of economic development and population growth has resulted in an increasing demand for transport, creating excessive pressure on the city transport infrastructure} & Hangzhou (China):  Bicycles and electric scooters are very popular , and major streets have dedicated bike lanes throughout the city & Nizhny Novgorod (Russia): \textit{Together with the metro it forms a system of high - speed rail transport of the city}\\
    \hline
    \end{tabular}
    \label{tab:best_similar_text_examples}
\end{table}

\begin{figure}[!htb]
\begin{center}
\includegraphics[scale=.49]{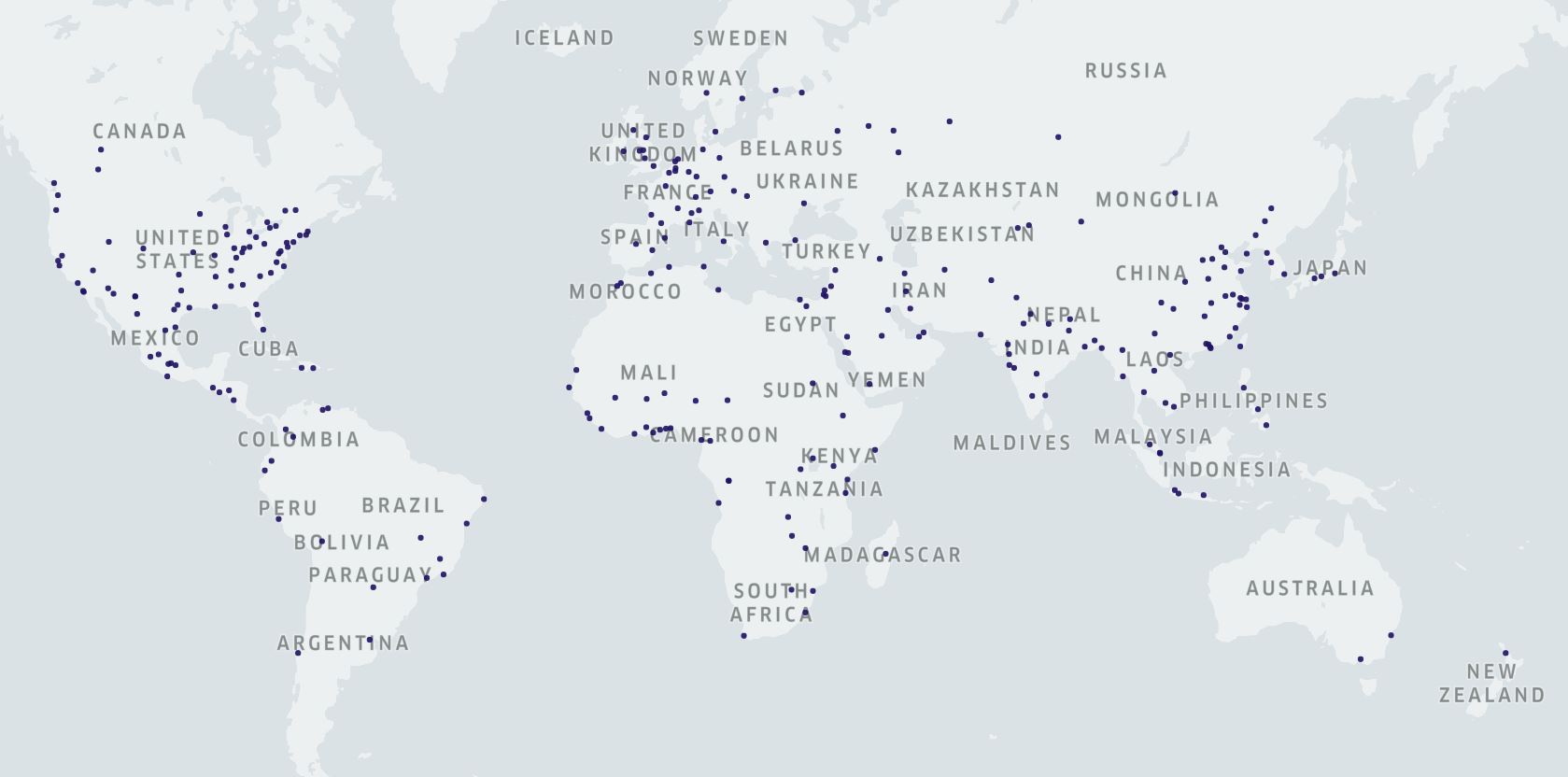}
\caption{Cities selected from \cite{okeMIT} for developing the typology classification models in our study} \label{fig:mit_train_test}
\end{center}
\end{figure}

\begin{figure}[!htb]
\begin{center}
\includegraphics[scale=.49]{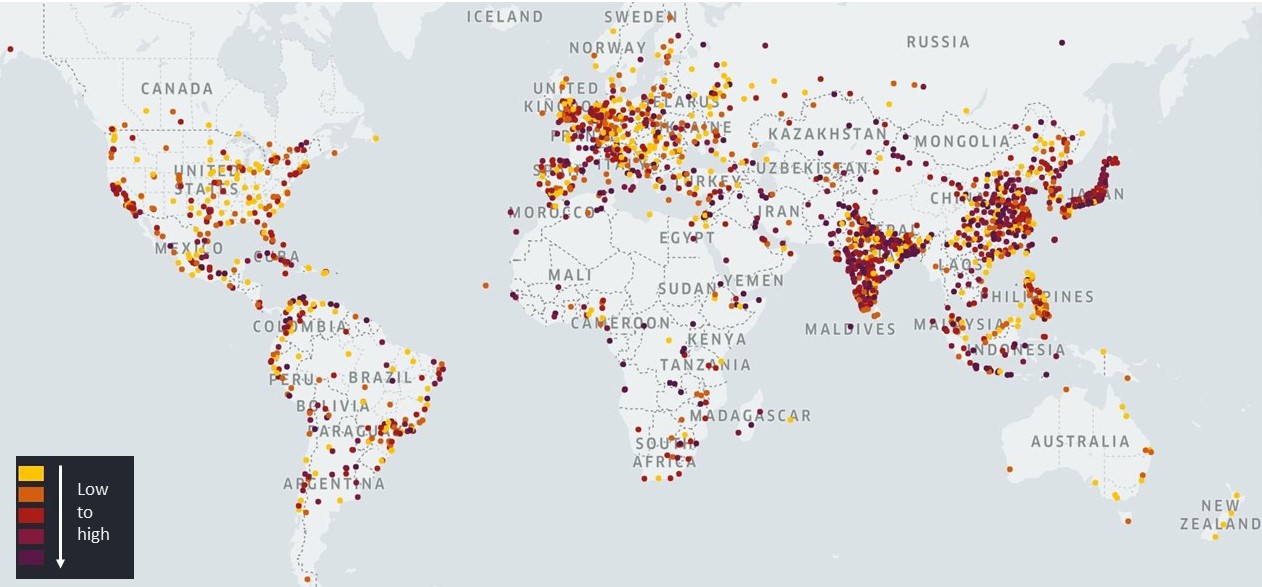}
\caption{2,102 cities from Wikipedia and their congestion probability scores, the sequential color palette represent low to high range of probability values. Higher values indicate that there is a high chance that the city is congested and vice-versa.} \label{fig:wikicities_plot}
\end{center}
\end{figure}

\paragraph*{Application of city typology models beyond test set} The models developed for city typology predictions in our study can be applied to any city in the world whose details are available on Wikipedia. For the purpose of building the typology classification models, we use the Wikipedia data for 282 cities (including train and test set).
Figure~\ref{fig:mit_train_test} shows the locations of these cities on the world map. 
To extend the analysis to other cities across the globe, we collect the list of cities (and urban towns) with 100,000 or more inhabitants from \cite{wiki_cities} (many other cities are available on Wikipedia based on different criteria). We select around 2102 cities by web crawling the Wikipedia list of cities pages, and fetch the city Wikipedia URLs. Using these URLs, for each city, we obtain the necessary data (following the same process as in Section~\ref{sec:data}). Based on the input variables used in the best performing typology classification models (refer Table~\ref{tab:model_coeff}), we compute the feature vector for these cities (as described in Section~\ref{sec:performance}). To provide a sense of how a city typology study on limited samples can be propagated to a larger scale using our proposed method, Figure~\ref{fig:wikicities_plot} shows the application of one of the typology classification models from our study. The figure includes $\sim$ 2,100 city locations (exclusive of the city data by \cite{okeMIT}) and the choropleth map of congestion probability scores obtained using the congestion classification model.\\

\paragraph*{Application of Wikipedia features beyond city typology label prediction}
To further test the relevance and effectiveness of our proposed method for extracting information from Wikipedia for downstream applications, we use the keyline features for a separate task (other than the city typology prediction). For this purpose, we consider the microtransit service feasibility prediction task. Microtransit service is a tech-enabled shared transportation that provides flexible routing (between  traditional fixed route transit and ride hailing technology) and/or highly flexible scheduling of minibus like vehicles. One example of such microtransit services available in various cities is Via Transportation. Via operates in more than 20 countries working in partnership with over 90 transit agencies, private companies, school districts, and universities; operations include first-last-mile connections, paratransit services, school networks and others. While Via is growing rapidly expanding its operations to multiple cities worldwide, the microtransit operator's decision of selecting a city among candidate cities for the growth of their services may be induced by the city characteristics along with various other factors. In this context, we hypothesize that cities with some patterns in the mobility-based features (\emph{i.e.}, $f^{(c)}, f^{(a)}, f^{(t)}, f^{(b)}$ considered in our study) may positively affect the chance of Via services being operated in that city. Although, there are factors beyond Wikipedia-based features that might dominate the decision of selection of new cities (for operation of such services). However, based on our hypothesis, the microtransit feasibility prediction exercise is to test the relevance of the Wikipedia features (computed using our proposed method) in a downstream application task other than city typology classification.

For this task, we compiled a list of $71$ Via cities worldwide (\emph{i.e.}, cities where Via provides its services as shown in Figure~\ref{fig:Viadeployments}); 33 out of these 71 cities are present in the typology dataset considered in our study (\emph{i.e.}, train and test set in Section~\ref{sec:data}). Based on the typology labels available for Via and non-Via cities (\emph{i.e.}, cities where Via does not operate currently), we first analyze if there is exists a pattern between city typologies and Via cities. This is done to examine (based on the typology data) if cities with a particular typology have a higher chance of being a Via city than cities which don’t belong to that typology. For example, in many cities Via operates in partnership with transit agencies, hence there is a possibility that there is a positive relation between transit-heavy cities and Via cities. 
To investigate such relations, we compute the probabilities of cities being a Via city given the city typologies; for auto-heavy typology this relation can be defined as follows:\\
\begin{align}
 \dfrac{\mathbb{P}(V|A)}{\mathbb{P}(V|A^{\sim})} &= \dfrac{\mathbb{P}\left( \text{City is a Via city} | \text{ city is auto-heavy}  \right)}{\mathbb{P}\left( \text{City is a Via city} | \text{ city is not auto-heavy}  \right)}\label{eq:Via_autotypology_prob}
\end{align}
Using Bayes' theorem (\citealp{bayes1991essay}), \cref{eq:Via_autotypology_prob} can be written as
\begin{align}
 \dfrac{\mathbb{P}(V|A)}{\mathbb{P}(V|A^{\sim})} &= \dfrac{\dfrac{\mathbb{P}(A|V).\mathbb{P}(V)}{\mathbb{P}(A)}}{\dfrac{\mathbb{P}(A^{\sim}|V).\mathbb{P}(V)}{\mathbb{P}(A^{\sim})}} =  \left(\dfrac{\mathbb{P}(A|V)}{1-\mathbb{P}(A|V)} \right) \left(\dfrac{1-\mathbb{P}(A|V)}{1-\mathbb{P}(A)}\right)
 %\left( \dfrac{1-\mathbb{P}(A)}{\mathbb{P}(A)} \right) \nonumber\\
  % &= \dfrac{\mathbb{P}(V)\left(\dfrac{\mathbb{P}(A|V)}{\mathbb{P}(A)}\right)}
  %{\mathbb{P}(V)\left(\dfrac{1-\mathbb{P}(A|V)}{1-\mathbb{P}(A)}\right)}\nonumber\\
 % &= \dfrac{\left(\dfrac{\mathbb{P}(A|V)}{\mathbb{P}(A)}\right)}
  %{\left(\dfrac{1-\mathbb{P}(A|V)}{1-\mathbb{P}(A)}\right)}
 %&=\left( \dfrac{\mathbb{P}(A|V)}{1-\mathbb{P}(A|V)} \right) \left( \dfrac{1-\mathbb{P}(A)}{\mathbb{P}(A)} \right) 
\end{align}

Based on the cities in the typology data %(\emph{i.e.}, 197 cities in 
(train set only) along with the details of Via services in these cities (\emph{i.e.}, Via or non-Via city) and auto-heavy typology labels of the cities, the value of the above expression is calculated as 2.86. This indicates that if a city is auto-heavy, the chances of it being a Via city is 2.86 times higher compared to cities that are not auto-heavy. Similar calculations for transit-heavy typology show that transit-heavy cities are 2.1 times more likely to be a Via city than cities that are not transit-heavy. For congestion and bike-friendly typology, these values are found to be less than 1.

Considering that there exists some pattern between city typologies and Via microtransit services in cities (based on the above derived relations), we focus on the task of predicting the feasibility of such microtransit services in cities given typology-based city features. For this task, we use a new source of ground truth data based on the list of Via operations in cities (\emph{i.e.}, 71 cities as mentioned above). In particular, we build a logistic regression model with Via and non-Via cities (1s and 0s) using congestion, auto, transit, bike keyline features ($f_{opt}^{(c)}$, $f_{opt}^{(a)}$, $f_{opt}^{(t)}$, $f_{opt}^{(b)}$ computed based on optimal feature keyline sets as obtained in Section~\ref{sec:keyline set expansion results}), and population density ($f^{(density)}$) from Wikipedia infobox as input variables for cities. This model is tasked to predict the probability estimate of a city being a Via city (indicating microtransit feasibility in the city). We use the city typology dataset (282 cities which includes 33 Via cities) along with additional Via cities (38 cities) for this model.
%The model features $f_{opt}^{(c)}$, $f_{opt}^{(a)}$, $f_{opt}^{(t)}$, and $f_{opt}^{(b)}$ for Via cities are computed using optimal feature keyline sets (as in Section~\ref{sec:keyline set expansion results}). 
This provides us in total $320$ cities with a 4-dimensional feature vector for each city, where cities with Via services (Via cities) were labeled as $1$ and others (non-Via cities) as $0$. We compute the input feature values for these cities and train the LR model (for microtransit feasibility prediction) using 70\% of the data, while 30\% data is used for testing the model performance. Using the trained LR model on the test set, we obtain AUC scores of 0.65 for train set and 0.62 for test set (this is much better than 0.5 which would be the AUC from just randomly guessing the label). The estimated coefficient values of the model %\textit{intercept}, $f_{opt}^{(c)}$, $f_{opt}^{(a)}$, $f_{opt}^{(t)}$, $f_{opt}^{(b)}$, and $f^{(density)}$ 
are summarized in Table~\ref{tab:microtransit_feasibility_model}.

%%% MICROTRANSIT FEASIBILITY MODEL OUTPUT %%%
\begin{table}[h]
\centering
    \caption{Microtransit feasibility prediction model results using Wikipedia data}
    \scalebox{0.98}{
    \begin{tabular}{c| c| c| c| c| c| c| c}
    \hline
    \multicolumn{6}{c|}{Model features} & \multicolumn{2}{c}{Model scores}\\
    \hline
	Intercept	&	$f_{opt}^{(c)}$	&	 $f_{opt}^{(a)}$	&	$f_{opt}^{(t)}$	& $f_{opt}^{(b)}$	&	$f^{(density)}$	&	AUC train	&	AUC test \\
%\\[-0.8em]
     \hline
     %\\[-0.8em]
	-0.687	&	-2.489	& 0.223	& 0.294	& 2.74	&	-2.385 &	0.65	&	0.62\\
    \hline
    \end{tabular}}
    \label{tab:microtransit_feasibility_model}
\end{table}
%%%%%%%%%%%%%%%%%%%%%%%%%%%%%%%%%%%%%%%%%%
%\begin{table}[h]
%\centering
 %   \caption{Microtransit feasibility prediction model results using Wikipedia data}
 %   \scalebox{0.98}{
 %   \begin{tabular}{|c c|}
  %  \hline
  %  Model & Values\\
  %  \hline
  %  Intercept & -0.687\\
  %  \hline
  %  $f_{opt}^{(c)}$	& -2.489\\
  %  \hline
  %  $f_{opt}^{(a)}$	& 0.223\\
  %  \hline
  %  $f_{opt}^{(t)}$	& 0.294 \\
  %  \hline
  %   $f_{opt}^{(b)}$ & 2.74\\
  %   \hline
  %  $f^{(density)}$	&	-2.385\\
  %  \hline
  %  AUC train	&	0.65\\
  %  \hline
  %  AUC test & 0.62\\
  %  \hline
  %  \end{tabular}}
  %  \label{tab:microtransit_feasibility_model}
%\end{table}

%-0.687, -2.489, 0.223,  0.294,  2.74, and -2.385 respectively. 
It is particularly interesting to note the positive coefficients for auto and transit features ($f_{opt}^{(a)}$ and $f_{opt}^{(t)}$) considering the patterns found earlier (using \cref{eq:Via_autotypology_prob}) for Via cities with auto-heavy and transit-heavy typologies. 
This experiment suggests that the information extracted from Wikipedia can be useful for mobility-based prediction tasks and can be generalized to other city level classification tasks as well. In the latter case, Wikipedia data may very well be used as a complementary data source. 

\textcolor{black}{Additionally, our proposed method can be applied to other text-based data sources (\emph{e.g.}, online articles on cities, city agency websites and others). Coupled with other data sources, city features from Wikipedia can be used in various city-level analyses and forecast models by transportation agencies and urban planners. For instance, this could benefit the ongoing efforts and research on decarbonization of cities across multiple countries (\citealp{doe}) such as cost-benefit analysis of different transport policies (e.g., congestion charge) as climate change mitigation measures in different cities (\citealp{creutzig2012decarbonizing}). 
Another interesting application of the data extracted from Wikipedia using the proposed approach could include perception analysis of users (city residents) with regards to the transportation/mobility aspects of cities. Hence, for various data driven large-scale city-level analyses, our proposed method can serve as a useful tool for effectively utilizing new text-based data sources to aid subsequent transportation and urban planning efforts.}

%% CONCLUSION %%
\section{Conclusion} \label{sec:conclusion}
In this study, we propose a novel method for the utilization of Wikipedia articles on cities for a large-scale global city typology prediction (focusing on the transportation aspect of cities). Using data extracted from Wikipedia, we develop four typology classification models to predict congestion, auto-heavy, transit-heavy, and bike-friendly type cities (based on ground truth labels by \citealp{okeMIT}). We do so by algorithmically extracting lines from a city’s Wikipedia page which semantically match a typology (via the SBERT NLP model), and use the typology-wise match scores to derive a low dimensional keyline-based feature vector (4 dimensional) representing the city. Using our proposed method, we demonstrate how a city typology classification based on limited samples ($\sim$ 300 cities) can be proliferated to an enormous scope spanning over 2000 cities across the world. \textcolor{black}{To foster further developments in this direction as well for use in other downstream tasks by the transportation research community, we have made the dataset (on typology inferences and city (typology-based) features) of 2102 worldwide cities publicly accessible via Zenodo (\citealp{zenodo})}.

Wikipedia contains unprecedented volume and variety of information on different cities across the globe. Wikipedia not only contains detailed information on various aspects of cities, it also offers information on the intangible aspects such as people's preferences and opinions (since the information is user-generated); the way people express these details vary based on locations, regions, and countries. In order to capture such variety of information pertaining to a specific typology, we present an iterative keyline expansion method that selects a representative set of keylines from city Wikipedia pages that allude to the city typology. The optimal feature keyline sets obtained using our proposed algorithm can be used to identify similar lines indicative of respective typologies for any city in the world that has a Wikipedia page with relevant transportation information in it. 

Our methodology permits integration of information from textual and numeric components (on Wikipedia) in the typology prediction models, and provides sufficient flexibility for expansion of the model feature vector allowing incorporation of additional variables. As our approach is based on understanding the intuitive meaning of the typology to consequently extract semantically similar and relevant textual information for defining city characteristics, it can be easily extended to different typology prediction tasks. As long as there is sufficient information pertaining to such typologies available in city Wikipedia pages, this holds true for both transportation-based typologies (\emph{e.g.}, pedestrian-friendly or walkable cities, paratransit accommodating cities) and non-transportation-based ones as well (\emph{e.g.}, climate-friendly cities). Additionally, the keyline-based features from Wikipedia can possibly complement other downstream application tasks like the one discussed in our study for microtransit feasibility prediction.

Our study finds Wikipedia articles to be informative about transportation-based typology indicators. To the best of our knowledge, this is the first time the text-based information from Wikipedia articles are used as a data source for cities in this manner. This opens up new opportunities for utilizing text-based data for transportation studies. Improving the relevance of the proposed keyline selection algorithm by understanding the nature of noisy selected lines
%studying their difference with already selected lines in previous iterations, 
and introducing penalties for noise to make it more robust are promising future research directions.   \textcolor{black}{Along those lines, determining top-$k$ keylines out of the $n$ optimal keylines per typology with each keyline (similarity score) as a model feature can potentially be an interesting direction for future study.}
%As a future research direction, it may be promising to improve the relevance of keyline sets by using explicitly detecting noisy lines the keyline sets can be additionally improved to ensure more robust results by penalizing introduction of noise in the data extracted from Wikipedia.
Another direction for future research could be around using additional data from Wikipedia (\emph{i.e.}, economic, social, government, and other factors) or from different text-based sources (such as online news and blogs on cities, official city websites, transit agency web pages) to supplement the city keyline features for predicting city typologies. The acquisition cost of obtaining necessary data from Wikipedia and the availability of advanced NLP models (like SBERT) is an important aspect that makes Wikipedia an appealing source of data for our study. However, we do recognize the need to ensure information quality in utilizing Wikipedia data for downstream applications. Although an increasing number of contributors and editors refining the information in Wikipedia makes this concern less severe, metrics to estimate to what extent Wikipedia data is representative of the reality can be developed to ensure better information quality. 

%In particular, extracting useful transportation related information from city Wikipedia pages as proposed in our study can be a valuable tool to complement and augment urban planning strategies in various contexts. 
Our novel approach of using text-based information from Wikipedia for understanding city typologies, and the outcomes presented in our study can assist a diverse group of stakeholders in transportation and urban planning fields. Additionally, we believe our method will reinforce existing studies utilizing crowd-sourced data leading to advances in strategic urban and transportation planning particularly in data-scarce regions. \textcolor{black}{Due to challenges in obtaining data with typology information on cities with data-scarcity (i.e., lack of open databases for public use), validating the proposed approach on such cities is difficult. In this context, based on the observed performances of the proposed models on cities beyond the training set and the fact that Wikipedia information is available even for cities that lack other data sources, we believe that Wikipedia can be used as an alternative, low-cost, and effective data source for cities facilitating transportation research.}
%Upon acceptance of this submission, we plan on releasing the dataset of over 2,000 cities in Wikipedia with their keyline features as a resource for the transportation research community to foster further developments in this direction.

\section*{Acknowledgement}
\noindent
The authors wish to acknowledge the funding support from C2SMART University Transportation Center (USDOT \#69A3551747124).
%thank the C2SMART University Transportation Center at NYU for its support for their research project.
%and the set of expanded keylines 

\bibliography{references} 
\end{document}